\pdfoutput=1
\documentclass[11pt]{article}

\usepackage[final]{acl}

\usepackage{times}
\usepackage{latexsym}

\usepackage{xcolor}
\usepackage{arydshln} 
\usepackage{booktabs}

\usepackage{graphicx}
\usepackage{tikz}
\usepackage{array}
\usepackage[edges]{forest}

\usepackage[T1]{fontenc}
\usepackage[utf8]{inputenc}

\usepackage{microtype}

\usepackage{inconsolata}

\usepackage{graphicx}

\usepackage{booktabs}
\usepackage{array}
\usepackage{pifont}
\usepackage{wasysym}
\usepackage{amssymb}
\usepackage{marvosym}
\usepackage{multirow}
\usepackage{graphicx}

\newif\ifshowcomments
\showcommentstrue 
\title{Code to Think, Think to Code: A Survey on Code-Enhanced Reasoning and Reasoning-Driven Code Intelligence in LLMs}

\author{
  \textbf{Dayu Yang\textsuperscript{\Scorpio}}\thanks{Equal contribution.} 
  \quad 
  \textbf{Tianyang Liu\textsuperscript{\Neptune}}\footnotemark[1] 
  \quad 
  \textbf{Daoan Zhang\textsuperscript{\twonotes}}\footnotemark[1]
  \\
  \textbf{Antoine Simoulin\textsuperscript{\Scorpio}} 
  \quad 
  \textbf{Xiaoyi Liu\textsuperscript{\Scorpio}} 
  \quad 
  \textbf{Yuwei Cao\textsuperscript{\Scorpio}} 
  \quad
  \textbf{Zhaopu Teng\textsuperscript{\Scorpio}}
  \quad
  \textbf{Xin Qian\textsuperscript{\Scorpio}}
  \\
  \textbf{Grey Yang\textsuperscript{\Scorpio}}\thanks{Equal advising.}
  \quad 
  \textbf{Jiebo Luo\textsuperscript{\twonotes}}\footnotemark[2]
  \quad
  \textbf{Julian McAuley\textsuperscript{\Neptune}}\footnotemark[2] 
  \\
  \textsuperscript{\Scorpio}Meta AI 
  \quad
  \textsuperscript{\Neptune}University of California, San Diego 
  \quad \textsuperscript{\twonotes}University of Rochester 
  \\
  \small{\texttt{\{dayuyang,antoinesimoulin,xiaoyiliu,yuweicao,zhaoputeng,xinqian,glyang\}@meta.com}} 
  \\
  \small{\texttt{\{til040,jmcauley\}@ucsd.edu}} 
  \quad 
  \small{\texttt{\{daoan.zhang,jluo\}@rochester.edu}}
}

\begin{document}
\maketitle

\begin{abstract}

In large language models (LLMs), code and reasoning reinforce each other: code offers an abstract, modular, and logic-driven structure that supports reasoning, while reasoning translates high-level goals into smaller, executable steps that drive more advanced code intelligence.
In this study, we examine how code serves as a structured medium for enhancing reasoning: it provides verifiable execution paths, enforces logical decomposition, and enables runtime validation. We also explore how improvements in reasoning have transformed code intelligence from basic completion to advanced capabilities, enabling models to address complex software engineering tasks through planning and debugging. Finally, we identify key challenges and propose future research directions to strengthen this synergy, ultimately improving LLM's performance in both areas.
% In this study, we examine how code serves as a structured medium for enhancing reasoning - providing verifiable execution paths, enforcing logical decomposition, and enabling runtime validation, and how advances in reasoning have transformed code intelligence from basic completion to sophisticated agent - enabling models to tackle complex software engineering tasks through deliberate planning and systematic debugging. Finally, we identify key challenges and propose future research directions may deepen the synergy, ultimately advancing LLM performance in both complex reasoning and code intelligence.

% My thoughts:
% We exaime how code serves as a structured medium for enhancing reasoning - providing verifiable execution paths, enforcing logical decomposition, and enabling runtime validation, and how advances in reasoning have transformed code intelligence from basic completion to sophisticated agent - enabling models to tackle complex software engineering tasks through deliberate planning and systematic debugging.

%recently, reasong/code

% gap/ attract

% in this work, 

\end{abstract}

\section{Introduction}
\label{sec:introduction}

\begin{figure}[t]
    \centering
    \includegraphics[trim={4.8cm 1cm 2cm 0.1cm},clip,width=1\linewidth]{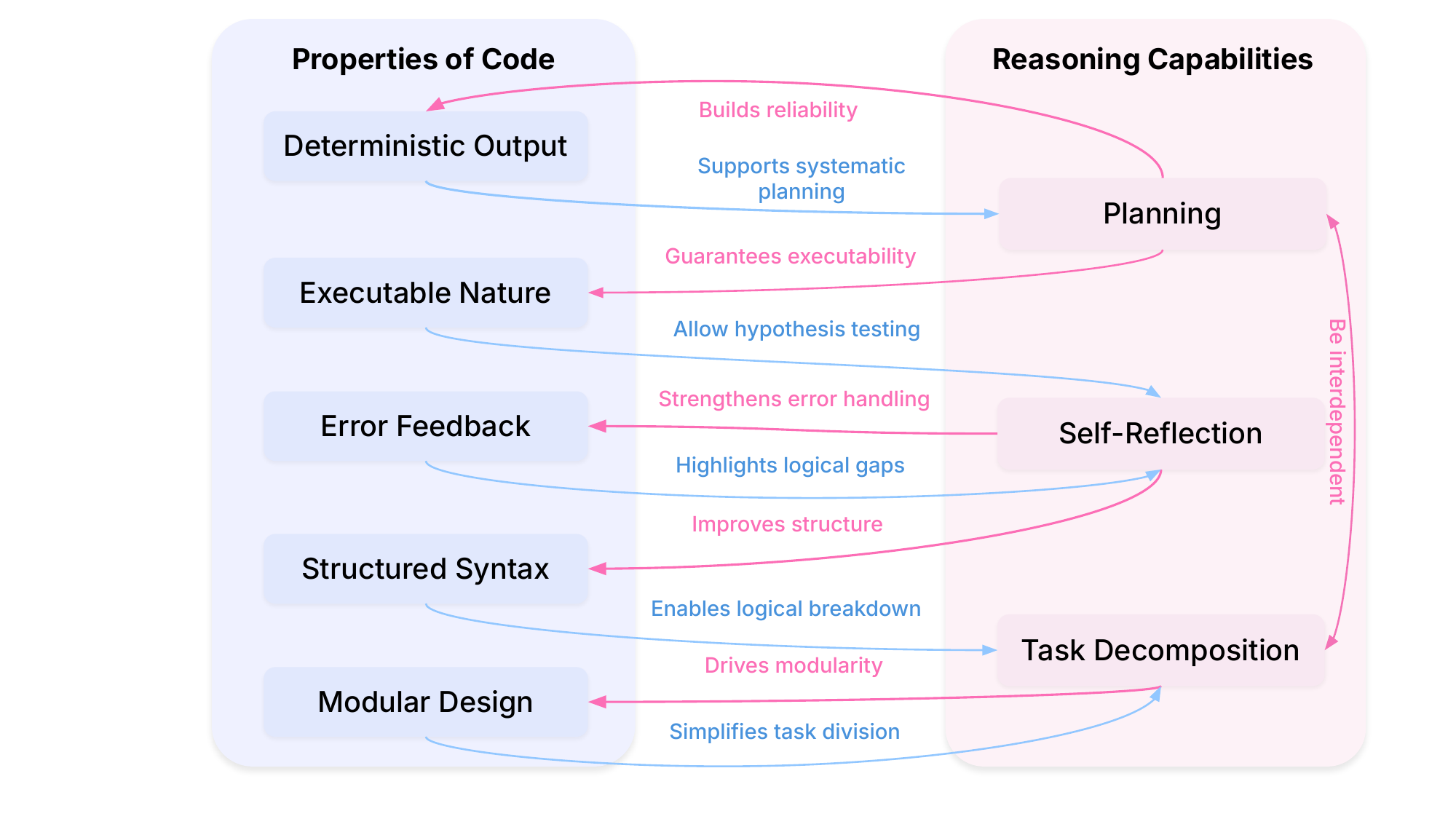}
    \caption{Bidirectional enhancement between code properties and reasoning capabilities.}
    \label{fig:synergy}
    \vspace{-2mm}
\end{figure}

Researchers have observed an intriguing ``Möbius strip'' effect: learning programming strengthens students' ability to solve complex problems, while strong analytical skills in turn speed up programming learning~\cite{brito2019reasoning}. This virtuous cycle now appears in artificial intelligence: When LLMs acquire code capabilities, they not only become more proficient programmers but also demonstrate significantly enhanced reasoning abilities across diverse domains such as mathematical deduction and logical inference. As their reasoning capacity evolves, these systems increasingly tackle complex programming challenges, even showing potential to outpace human developers~\cite{chowdhury2024swebenchverified}. Recent breakthrough models like OpenAI-o1~\cite{openai2024openaio1card} and DeepSeek-R1~\cite{guo2025deepseek} show powerful task-solving capabilities, particularly advances in reasoning. A key factor driving this transformation has been the strategic integration of code - both during pre-training phases~\cite{touvron2023llama} and reasoning processes~\cite{chen2022program}. The rigorous logical structure of code provides a unique ``training ground'' for strengthening LLMs' reasoning capabilities, while AI's evolving reasoning abilities continuously enhance code intelligence. This bidirectional relationship reveals profound intrinsic connections between coding and reasoning (see Figure~\ref{fig:synergy}).

In this bidirectional enhancement process, core properties of code - including structured syntax, execution feedback, and modular design - significantly promote task decomposition, reasoning chain construction, and self-reflection (\S\ref{sec:c4r}). Conversely, improved reasoning capabilities drive advances in code intelligence, such as task decomposition, code comprehension and modification, program debugging and optimization, ultimately giving rise to intelligent agents capable of end-to-end software development (\S\ref{sec:irc}, \S\ref{sec:agent}). For instance, advanced reasoning techniques like Chain-of-Thought prompting~\cite{wei2022chain, zhang2024cocot} and Self-Reflection~\cite{shinn2024reflexion} are expanding code generation from simple autocompletion to intelligent software development assistants~\cite{devinAI, yang2024if}, even capable of managing complete software engineering lifecycles~\cite{jimenez2024swebenchlanguagemodelsresolve}.

Despite these promising strides, there has been limited systematic review of how code and reasoning interact and reinforce each other. To address this gap and provide a structured view of the code-reasoning synergy in LLMs, we pose the following core questions:
(1) \textbf{How do code representations influence LLM reasoning?}
(2) \textbf{How do advances in LLM reasoning reshape code intelligence systems?}
(3) \textbf{What challenges arise from the code reasoning interplay in LLMs?}
% Answers to these questions are explored through a structured analysis of the code-reasoning synergy in the following sections. We first investigate how code representations contribute to LLM reasoning, particularly in inference and training, examining their role in improving logical consistency, enhancing precision, and strengthening reasoning capabilities (Sec.\ref{sec:cer}). Next, we explore how advancements in LLM reasoning reshape code intelligence, enabling a shift from static pattern matching to an adaptive, iterative workflow that integrates planning, debugging, and self-optimization (Sec.\ref{sec:rec}). Finally, we highlight key challenges in the code-reasoning synergy, including limitations in commonsense reasoning, cross-linguistic adaptability, and large-scale code comprehension, which necessitate innovations in dataset curation, hybrid training strategies, and evaluation frameworks (Sec.~\ref{sec:fut}).
% By systematically examining these dimensions, this survey synthesizes existing progress and identifies critical areas for future research, paving the way for deeper integration between reasoning and code intelligence.

To systematically investigate these questions, our research unfolds along the following dimensions: (i) analyzing how code serves as an effective reasoning medium, helping LLMs structure their reasoning and validate results (\S\ref{sec:cer}); (ii) exploring how enhanced reasoning capabilities expand the boundaries of code intelligence (\S\ref{sec:rec}); and (iii) summarizing current challenges, focusing on open problems in model interpretability, scalable training, and multimodal fusion, while proposing future research directions (\S\ref{sec:fut}).

\definecolor{hidden-pink}{RGB}{255,245,247}
\definecolor{hidden-draw}{RGB}{107,114,128}

\definecolor{level-1}{RGB}{240,249,255}  % Light blue
\definecolor{level-2}{RGB}{236,253,245}  % Light green
\definecolor{level-3}{RGB}{254,242,242}  % Light red

\tikzstyle{my-box}=[
    rectangle,
    draw=hidden-draw,
    rounded corners,
    text opacity=1,
    minimum height=2em,
    minimum width=5em,
    inner sep=2pt,
    align=center,
    fill opacity=.5,
    line width=0.8pt,
]
\tikzstyle{citation}=[my-box, minimum height=2em,
    fill=hidden-pink!80, text=black, align=left,font=\normalsize,
    inner xsep=2pt,
    inner ysep=5pt,
    line width=0.8pt,
]
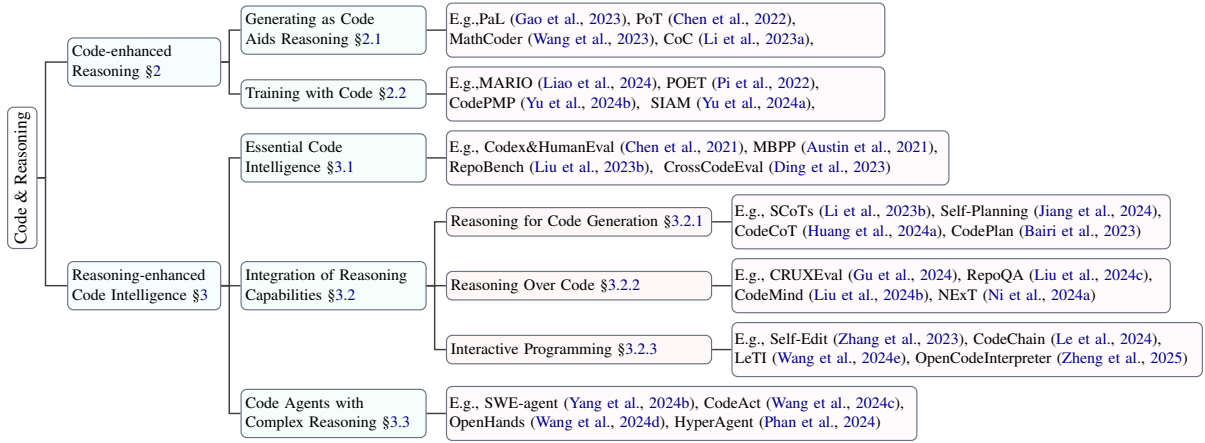
\begin{figure*}[!th]
    \centering
    \resizebox{0.99\textwidth}{!}{
        \begin{forest}
            forked edges,
            for tree={
                grow=east,
                reversed=true,
                anchor=base west,
                parent anchor=east,
                child anchor=west,
                base=left,
                font=\large,
                rectangle,
                draw=hidden-draw,
                rounded corners,
                align=left,
                minimum width=6em,
                edge+={darkgray, line width=1pt},
                s sep=6pt,
                inner xsep=3pt,
                inner ysep=3pt,
                line width=0.8pt,
                ver/.style={rotate=90, child anchor=north, parent anchor=south, anchor=center},
            },
            where level=1{
                text width=9.0em,
                font=\normalsize,
                fill=level-1,
                fill opacity=.5,
            }{},
            where level=2{
                text width=11.0em,
                font=\normalsize,
                fill=level-2,
                fill opacity=.5,
            }{},
            where level=3{
                text width=16.0em,
                font=\normalsize,
                fill=level-3,
                fill opacity=.5,
            }{},
            where level=4{
                text width=18.0em,
                font=\normalsize,
                fill=hidden-pink,
                fill opacity=.5,
            }{}
            [
                Code \& Reasoning , ver
                [                
                    Code-enhanced \\Reasoning \S\ref{sec:cer}
                    [
                        Generating as Code\\Aids Reasoning \S\ref{sec:c4r}
                        [
                            {E.g.,}PaL~\cite{gao2023pal}{, }PoT~\cite{chen2022program}{, }\\MathCoder~\cite{wang2023mathcoder}{, }CoC~\cite{li2023chain}{, }
                            , citation, text width=27em
                        ]
                    ]
                    [
                        Training with Code \S\ref{sec:twc}
                        [
                            {E.g.,}MARIO~\cite{liao2024mario}{, }POET~\cite{pi2022reasoning}{, }\\
                            CodePMP~\cite{yu2024codepmp}{, }
                            SIAM~\cite{yu2024siam}{, }
                            , citation, text width=27em
                        ]
                    ]
                ]
                [
                    Reasoning-enhanced \\Code Intelligence \S\ref{sec:rec}
                    [
                        Essential Code \\ Intelligence \S\ref{sec:eci}
                        [
                            {E.g., }Codex\&HumanEval~\cite{chen2021evaluatinglargelanguagemodels}{, }MBPP~\cite{austin2021programsynthesislargelanguage}{, }\\
                            RepoBench~\cite{liu2023repobenchbenchmarkingrepositorylevelcode}{, } CrossCodeEval~\cite{ding2023crosscodeevaldiversemultilingualbenchmark}
                            , citation, text width=31em
                        ]
                    ]
                    [
                        Integration of Reasoning \\Capabilities \S\ref{sec:irc}
                        [
                            Reasoning for Code Generation \S\ref{sec:r4c}
                            [
                                {E.g., }SCoTs~\cite{li2023structuredchainofthoughtpromptingcode}{, }Self-Planning~\cite{jiang2024selfplanningcodegenerationlarge}{, }\\
                                CodeCoT~\cite{huang2024codecottacklingcodesyntax}{, }CodePlan~\cite{bairi2023codeplanrepositorylevelcodingusing}
                                , citation, text width=27em
                            ]
                        ]
                        [
                            Reasoning Over Code \S\ref{sec:roc}
                            [
                                {E.g., }CRUXEval~\cite{gu2024cruxevalbenchmarkcodereasoning}{, }RepoQA~\cite{liu2024repoqaevaluatinglongcontext}{, }\\
                                CodeMind~\cite{liu2024codemindframeworkchallengelarge}{, }NExT~\cite{ni2024nextteachinglargelanguage}
                                , citation, text width=27em
                            ]
                        ]
                        [
                            Interactive Programming \S\ref{sec:ip}
                            [
                                {E.g., }Self-Edit~\cite{zhang2023selfeditfaultawarecodeeditor}{, }CodeChain~\cite{le2024codechainmodularcodegeneration}{, }\\
                                LeTI~\cite{wang2024letilearninggeneratetextual}{, }OpenCodeInterpreter~\cite{zheng2025opencodeinterpreterintegratingcodegeneration}
                                , citation, text width=29em
                            ]
                        ]
                    ]
                    [
                        Code Agents with \\Complex Reasoning \S\ref{sec:agent}
                        [
                            {E.g., }SWE-agent~\cite{yang2024sweagentagentcomputerinterfacesenable}{, }CodeAct~\cite{wang2024executablecodeactionselicit}{, }\\
                            OpenHands~\cite{openhands2024}{, }HyperAgent~\cite{phan2024hyperagentgeneralistsoftwareengineering}
                            , citation, text width=29em
                        ]
                    ]
                ]
            ]
        \end{forest}
    }
    \caption{Taxonomy of interplay between Code and Reasoning.}
    \label{fig:taxonomy}
    \vspace{-3mm}
\end{figure*}

\begin{table*}[ht!]
\centering
\resizebox{\textwidth}{!}{
\begin{tabular}{@{}l l l c c c@{}}
\toprule
\textbf{Method Type} & \textbf{Method} & \textbf{Model} & \textbf{GSM8K} & \textbf{SVAMP} & \textbf{MATH} \\
\midrule

% Baselines
\multirow{2}{*}{\textbf{Baseline}} 
& Direct$^\dagger$ & Codex & 19.7 & 69.9 & -- \\
& CoT$^\dagger$~\cite{wei2022chain} & GPT-4 & 92.0 & 97.0 & -- \\
\midrule

% Single Execution
\multirow{2}{*}{\textbf{Single Execution}} 
& PAL~\cite{kabra2023program} & Codex & 72.0 & 79.4 & -- \\
& PoT~\cite{chen2022program} & GPT-4 & 97.2 & 97.4 & -- \\
\midrule

% Dynamic Code-Language Integration
\multirow{5}{*}{\textbf{Dynamic Code-Language}} 
& MathCoder-L~\cite{wang2023mathcoder} & Llama-2-70B & 83.9 & 84.9 & 45.1 \\
& MathCoder-CL~\cite{lu2024mathcoder2} & CodeLlama-34B & 81.7 & 82.5 & 45.2 \\
& CodePlan~\cite{wen2024unlocking} & Mistral-7B & 59.5 & 61.4 & 34.3 \\
& INC-Math~\cite{xiong2024inc} & GPT-4o-mini & -- & -- & 51.4 \\
\midrule

% Non-Executable Program Representations
\multirow{2}{*}{\textbf{Non-Executable}} 
& CoC~\cite{li2023chain} & text-davinci-003 & 71.0 & -- & -- \\
& CodePrompt~\cite{hu2023code} & GPT-3.5 (few-shot) & 80.6 & 79.6 & -- \\

\bottomrule
\end{tabular}}
\caption{Performance comparison of best-performing variants of code-aided reasoning methods across three key benchmarks (GSM8K, SVAMP, and MATH). Results show the percentage of problems solved correctly. ``--'' indicates no reported result. For each method, only the variant with highest GSM8K performance is shown (or highest MATH score when GSM8K is unavailable). $^\dagger$ "Direct" and "CoT" uses Codex model using few-shot direct prompting with/without CoT. The results are from~\citet{chen2022program}.}
\label{table:code-aided}
\vspace{-2mm}
\end{table*}

\section{Code-enhanced Reasoning}
\label{sec:cer}
\vspace{-2mm}
\subsection{Generating as Code Aids Reasoning}
\label{sec:c4r}
% intro
We examine how generating code and code-based training enhance LLMs' reasoning. By transforming reasoning problems into programmatic solutions, these approaches improve precision and reliability in complex reasoning tasks. The performance of major methods are listed in Table~\ref{table:code-aided}.

\subsubsection{Single Execution}
\label{subsec:single}

\citet{chen2022program, gao2023pal} introduced Program of Thoughts (PoT) and Program-aided language models (PaL), transforming numerical problem-solving into single-execution code generation tasks. Unlike chain-of-thought's natural language steps~\cite{wei2023chainofthoughtpromptingelicitsreasoning}, these approaches express the entire reasoning process as a self-contained executable program, providing a deterministic path to solutions while minimizing calculation errors.
\citet{bi2023when} investigated when this code-based transformation enhances reasoning, finding that PoT and PaL's effectiveness depends on code complexity. Their analysis revealed that code transformation benefits vary across problem types, suggesting generating code may not universally enhance reasoning.

% calibration
Recent work by~\citet{kabra2023program} examined how single-execution approaches affect model calibration, comparing self-assessment accuracy between code and language outputs. Their work demonstrated that code-based solutions often achieve better calibration than pure language.

\subsubsection{Dynamic Code-Language Integration}
\label{subset:dynamic}

% Dynamic involves code + text reasoning
Moving beyond pure code, \citet{wang2023mathcoder} explore a hybrid approach that weaves code snippets within explanatory text for reasoning. This integration was further refined by \citet{lu2024mathcoder2} through improved training data generation that pairs mathematical code with step-by-step reasoning.

% other than math, challenges in real-world problem -> needs planning

% planing : text + code generation 

When tackling real-world tasks with ambiguity, \citet{yang2024can, chen2024steering} identified LLMs' struggles in maintaining coherent reasoning when switching between code and language. To address this, \citet{wen2024unlocking} developed code-form plans to structure the integration, while \citet{xiong2024inc} established criteria for mode transitions.

% iteratively verify and execution...evolving process of generating text + code reasoning
Although these techniques improve code-language integration through control flow and mode-switching, they struggle with adapting to code execution errors. To address this, \citet{liu2024interactive} developed a REPL-based approach using execution results to guide natural language reasoning steps. Similarly, \citet{lei2024planning} created a workflow where natural language planning guides code generation, with execution feedback driving iterative refinements.

\subsubsection{Non-Executable Program Representations}

% not all problem can map to executable code space

Even when executable code is impractical for ambiguous or abstract reasoning problems, code-like structures can enhance reasoning processes. This has driven development of approaches that leverage code representations without requiring execution.

% chain of code.

Chain of Code~\citep{li2023chain} pioneered this direction by combining code snippets with undefined functions that are emulated through a "LMulator," enabling solutions for problems that mix computational and contextual reasoning. Building on this foundation, \citet{hu2023code} demonstrated that symbolic code generation serves as a reasoning scaffold, while Code Prompting~\citep{puerto2024code} reformulates problems into code-like templates to help models manage conditional logic. NExT~\citep{ni2024next} further advanced this approach by using execution traces to help models conceptually track program states without actual execution.

% also record states...more detailed simulation

% general frameowork using pseudo code to reasoning: "code form planning"

Recent work has expanded these approaches into broader frameworks. CodePlan~\citep{wen2024unlocking} introduces "code-form plans" --- structured pseudocode that organizes reasoning steps across diverse tasks. 
This systematization of non-executable program representations shows how code structures enhance reasoning coherence and interpretability without requiring execution. This approach has proven particularly valuable for problems that resist pure algorithmic solutions but benefit from programmatic organization~\citep{weir2024learning}, establishing non-executable program representations as a powerful tool for enhancing LLM reasoning.

\begin{table*}[ht!]
\centering
\resizebox{\textwidth}{!}{
\begin{tabular}{@{}c|llcccccc@{}}
\toprule
\textbf{Type} & \textbf{Model} & \textbf{Settings}& \textbf{Size} & \textbf{Metric} & & \textbf{Datasets} &  \\
\midrule

% commonsense reasoning
\multirow{3}{*}{\textbf{Math} }
& & & & & GSM8K & MATH & OCW \\ 
\cmidrule(lr){6-8}
& Lemma & Baseline & 34B & EM & 51.5 & 25.0 & 11.8 \\ % from Self-Planning paper
& \textbf{MARIO}~\cite{liao2024mario} & Proposed & 34B & EM & 78.2\textcolor{red}{(+26.7)} & 53.5\textcolor{red}{(+28.5)} & 30.2\textcolor{red}{(+18.4)} \\ % from Self-Collaboration paper
\midrule

\multirow{3}{*}{\shortstack{\textbf{Common Sense} \\ \textbf{Logic}}}
& & & & & HotpotQA & LogiQA & DROP \\ 
\cmidrule(lr){6-8}
& RoBERTa-L & Baseline & 355M & EM & 67.6 & 36.7 & 78.1 \\ % from Self-Planning paper
& \textbf{POET}~\cite{pi2022reasoning} & Proposed & 355M & EM & 68.7\textcolor{red}{(+1.1)} & 38.9\textcolor{red}{(+2.2)} & 79.8\textcolor{red}{(+1.7)} \\ % from Self-Collaboration paper
\midrule

\multirow{3}{*}{\shortstack{\textbf{Math} \\ \textbf{Logic}}}
& & & & &MathShepherd-pair& Reclor-pair& LogiQA2.0-pair \\ 
\cmidrule(lr){6-8}
& Qwen2-7B & Baseline & 7B & Reward & 0.88 &0.86& 0.83 \\ % from Self-Planning paper
& \textbf{CodePMP}~\cite{yu2024codepmp} & Proposed & 7B & Reward & 0.93\textcolor{red}{(+0.5)} & 0.87\textcolor{red}{(+0.1)} & 0.84\textcolor{red}{(+0.1)} \\ % from Self-Collaboration paper
\midrule

\multirow{3}{*}{\shortstack{\textbf{Math} \\ \textbf{Multi-lingual}}}
& & & & & APE& CMATH& GSM8K \\ 
\cmidrule(lr){6-8}
& Qwen2-Math & Baseline & 7B & Reward & 83.4 &87.3& 79.5 \\ % from Self-Planning paper
& \textbf{SIAM}~\cite{yu2024siam} & Proposed & 7B & Reward & 88.1\textcolor{red}{(+4.7)} & 93.2\textcolor{red}{(+5.9)} & 81.5\textcolor{red}{(+2.0)} \\ % from Self-Collaboration paper
\midrule

\multirow{3}{*}{\shortstack{\textbf{Instruction-Following} \\ \textbf{Decision-Making}}}
& & & & & AlpacaEval-2& MT-Bench& ALFWorld \\ 
\cmidrule(lr){6-8}
& Llama-2 & Baseline & 13B & Self-defined & 6.5 &6.1& 23.2 \\ % from Self-Planning paper
& \textbf{CODEPLAN}~\citep{wen2024unlocking} & Proposed & 13B & Self-defined & 12.2\textcolor{red}{(+5.7)} & 7.1\textcolor{red}{(+1.0)} & 33.3\textcolor{red}{(+10.1)} \\ % from Self-Collaboration paper

\bottomrule
\end{tabular}
}
\caption{Performance enhancement brought by training the model with code related data. "Baseline" denotes the vanilla model, while "Proposed" refers to the proposed methods.}
\label{table:code-training}
\vspace{-2mm}
\end{table*}

\subsection{Training with Code}
\label{sec:twc}

Code data strengthens LLMs' reasoning and planning abilities by providing structured patterns that guide logical thinking~\cite{touvron2023llama, achiam2023gpt, hu2024define}. This section examines how code data enhances these capabilities and discusses effective strategies for integrating code into LLM training.

\subsubsection{Empowering Reasoning and Planning Through Code Training}

Code-trained LLMs excel across various domains. In commonsense reasoning, \citet{madaan2022language} treats structured commonsense tasks as code generation problems, showing notable gains even when downstream tasks do not explicitly involve code. In mathematics, MathCoder~\cite{wang2023mathcoder} interleaves natural language, code, and execution results to improve mathematical reasoning. Its successor, MathCoder2~\cite{lu2024mathcoder2}, further refines these abilities with a higher-quality pre-training dataset that embeds mathematical reasoning steps in code.

Training on code also bolsters planning and decision-making. \citet{chen2024logic} used larger models to break down complex instructions into discrete functions, creating a function base for training smaller LLMs in structured planning. The dataset enables smaller models to acquire the planning and decision-making capabilities of their larger counterparts. Likewise, \citet{wen2024unlocking} curated a dataset of 2M standard prompt-response-code form plan triplets (prompt, response, code) to enhance models’ planning and decision-making.

% multi-modal 
In the multimodal domain, \textsc{ViStruct}~\citep{chen2023vistruct} utilizes the structure of programming it learned from code training to represent visual structural knowledge. This approach allows the model to capture structural information at different levels of granularity within images, enabling visual language models (VLMs) to better understand complex visual structures. This exemplifies how structured data, such as code, can serve as an excellent medium for visual data representation.

% real world
Code-trained LLMs and VLMs also shine in real-world scenarios. In multilingual environment settings, code acts as a bridge between languages.\cite{li2024eliciting} augments code datasets with machine-translated multilingual comments during training while preserving original code. Their approach uses step-by-step code primitives in prompts to derive facts and solutions, demonstrating code's effectiveness in multilingual reasoning.  In autonomous driving, \textsc{LaMPilot}~\citep{ma2024lampilot} achieves remarkable results by generating code based on user instructions and leveraging established functional primitives to replace ambiguous natural language commands. The approach showed exceptional results on the custom-built \textsc{LaMPilot Bench}. These applications highlight code data training's vast potential for reasoning and planning across real-world scenarios and environments.

\subsubsection{Training Strategies Based on Code}

Code-based LLMs have shown remarkable performance across domains. Here, we examine effective strategies for leveraging code data during model training to enhance their capabilities.

\paragraph{Code-only Training Strategies}

% method: data + post training
Incorporating code execution into traditional reasoning datasets boosts LLM performance. MARIO~\cite{liao2024mario} leverages both LLMs and human annotations to augments GSM8K~\cite{cobbe2021training} and MATH~\cite{hendrycks2021measuring} with Python interpreter traces, yielding significant downstream gains. Similarly, POET~\cite{pi2022reasoning} uses programs and execution results to train LLMs, showing improved natural language reasoning capabilities.
Furthermore, incorporating human preferences enhances training effectiveness~\cite{ding2024semcoder, zhang2024seppo}, CodePMP~\cite{yu2024codepmp} introduces a preference model pretraining pipeline using large-scale synthesized code-preference datasets, improving fine-tuning efficiency and reasoning performance. SIAM~\cite{yu2024siam} employs a code-based critic model to guide dataset construction through code generation and quality control, optimizing downstream performance.

\paragraph{Hybrid-data Training Strategies}
% data: strategy of code
Determining the optimal stage and proportion of code data in training LLMs is critical~\cite{tao2024crystal}. ~\citet{ma2023training} and~\citet{zhang2024unveiling} indicate that adding code during pretraining boosts general reasoning abilities, while adding code instructions during instruction tuning improves code-specific skills and adherence to human instructions. Mixing text and code data dynamically fosters progressive reasoning enhancements throughout training. Additionally, \citet{zhang2024unveiling} further finds that the effects of code data differ across reasoning domains but exhibit consistent trends within each domain. They conclude that optimal code mixing strategies are typically domain-specific rather than universal.

% \subsection{Program-Guided Reasoning}

% \subsection{Code-Augmented Training}

\begin{table*}[ht!]
\centering
\resizebox{\textwidth}{!}{
\begin{tabular}{@{}lllcccc@{}}
\toprule
\textbf{Method Type} & \textbf{Method} & \textbf{Model} & \textbf{HumanEval} & \textbf{MBPP} & \textbf{SWE-Bench (Lite)} \\
\midrule
% Baseline Methods
\multirow{2}{*}{\textbf{Baseline}} &
Direct$^\dagger$ & Codex & 48.1 & 49.8 & -- \\
& CoT$^\dagger$~\cite{wei2023chainofthoughtpromptingelicitsreasoning} & Codex & 53.9 & 54.5 & -- \\
\midrule
% Reasoning-enhanced Methods
\multirow{3}{*}{\textbf{Reasoning-enhanced}} &
SCoTs~\cite{li2023structuredchainofthoughtpromptingcode} & GPT-3.5 & 60.6 & 47.0 & -- \\
& Self-Planning~\cite{jiang2024selfplanningcodegenerationlarge} & Codex        &  60.3  & 55.7  & --   \\
& CodeCoT~\cite{huang2024codecottacklingcodesyntax} & GPT-3.5 & 79.3 & 89.5 & -- \\
\midrule
% Interactive Methods
\multirow{3}{*}{\textbf{Interactive}} &
Self-Edit$^\dagger$~\cite{zhang2023selfeditfaultawarecodeeditor} & GPT-3.5 & 62.2 & 52.4 & -- \\
& Self-Debugging~\cite{chen2023teachinglargelanguagemodels} & GPT-4 & -- & 80.6 & -- \\
& Self-Collaboration~\cite{dong2024selfcollaborationcodegenerationchatgpt} & GPT-3.5 & 74.4 & 68.2 & -- \\
& AgentCoder~\cite{huang2024agentcodermultiagentbasedcodegeneration} & GPT-4 & 96.3 & 91.8 & -- \\
\midrule
% Fine-tuned Methods
\multirow{2}{*}{\textbf{Fine-tuned}}
& CodeAct~\cite{wang2024executablecodeactionselicit} & Mistral-7B & 34.7 & 59.1 & -- \\
&OpenCodeInterpreter~\cite{zheng2025opencodeinterpreterintegratingcodegeneration} & DeepseekCoder-33B & 92.7 & 90.5 & -- \\
\midrule
% Agentic Methods
\multirow{4}{*}{\textbf{Agentic}} &
SWE-agent~\cite{yang2024sweagentagentcomputerinterfacesenable} & GPT-4 Turbo & -- & -- & 18.0 \\
& AutoCodeRover~\cite{zhang2024autocoderoverautonomousprogramimprovement} & GPT-4 & -- & -- & 19.0 \\
& OpenHands~\cite{openhands2024} & Claude-3.5-Sonnet & -- & -- & 26.0 \\
& HyperAgent~\cite{phan2024hyperagentgeneralistsoftwareengineering} & Claude-3.5-Sonnet & -- & -- & 26.0 \\
& Agentless$^\ddagger$~\cite{xia2024agentlessdemystifyingllmbasedsoftware} & GPT-4o & -- & -- & 27.3 \\
\bottomrule
\end{tabular}
}
\caption{Performance comparison of reasoning-enhanced code intelligence methods across benchmarks. Results reflect best performance from original papers except where noted ($^\dagger$results from Self-Planning~\cite{jiang2024selfplanningcodegenerationlarge} for Direct and CoT, and from CodeCoT~\cite{huang2024codecottacklingcodesyntax} for Self-Edit). $^\ddagger$Agentless represents an agent-free approach, while listed under Agentic methods for organization,  HumanEval and MBPP use pass@1 scoring, and ``--'' denotes unavailable or inapplicable results.}
\label{tab:r2c}
\vspace{-2mm}
\end{table*}

\section{Reasoning-Enhanced Code Intelligence}
\label{sec:rec}

Software development fundamentally requires intensive reasoning capabilities as developers decompose complex problems and rigorously analyze system behaviors and edge cases~\cite{hermans2021programmer}. Recent advances in LLMs have dramatically improved code generation capabilities~\cite{chen2021evaluatinglargelanguagemodels, rozière2024codellamaopenfoundation, li2023starcodersourceyou, codegemmateam2024codegemmaopencodemodels, deepseekai2024deepseekcoderv2breakingbarrierclosedsource, hui2024qwen25codertechnicalreport, li2022alphcode}, and their growing integration with reasoning capabilities has transformed code intelligence systems~\cite{austin2021programsynthesislargelanguage, yang2024sweagentagentcomputerinterfacesenable}. 
% This section explores how reasoning capabilities have revolutionized code intelligence, evolving tools from basic code generators to systems that can plan, reason about, and iteratively improve software solutions. We examine this progression through three stages: the limitations of direct code generation, the integration of explicit reasoning for code generation and comprehension, and the emergence of code agents for complex end-to-end development. 
This section examines the evolution of code intelligence through three stages: direct code generation's limitations, explicit reasoning integration for code generation and comprehension, and the emergence of code agents for complex end-to-end development. 
The performance of major methods are listed in Table~\ref{tab:r2c}.

\subsection{Essential Code Intelligence}
\label{sec:eci}
% This section mainly introduce code generation without reasoning
The foundation of modern code intelligence emerged with LLMs trained on code repositories, initially focusing on direct sequence prediction tasks like auto code completion, e.g., CodeXGLUE~\cite{lu2021codexgluemachinelearningbenchmark}, and docstring-based generation, e.g., HumanEval~\cite{chen2021evaluatinglargelanguagemodels} and MBPP~\cite{austin2021programsynthesislargelanguage}. These base models demonstrated capabilities in next-line prediction, fill-in-the-middle (FIM), and program synthesis~\cite{chen2021evaluatinglargelanguagemodels, xu2022systematicevaluationlargelanguage, bavarian2022efficienttraininglanguagemodels, fried2023incodergenerativemodelcode, li2023starcodersourceyou}, later extending to repository-level tasks like RepoBench~\cite{liu2023repobenchbenchmarkingrepositorylevelcode} and CrossCodeEval~\cite{ding2023crosscodeevaldiversemultilingualbenchmark}.
While these models excelled at simple tasks like code completion~\cite{githubcopilot}, their reliance on direct generation without explicit reasoning limited their effectiveness in complex scenarios requiring careful consideration of algorithmic design and edge case handling, or real-world programming scenarios that demand systematic planning.
% However, these early models lacked explicit reasoning steps in code generation, relying instead on direct generation from training data. While effective for simple tasks like direct code completion~\cite{githubcopilot}, they often struggled with complex problems requiring careful consideration of algorithmic design and edge case handling, or real-world programming scenarios that demand systematic planning.

\subsection{Integration of Reasoning Capabilities}
\label{sec:irc}
Modern models typically exhibit two key reasoning types when working with code: \textit{reasoning to code}, which involves planning and problem decomposition prior to implementation, and \textit{reasoning over code}, which focuses on understanding code behavior and properties. These reasoning forms naturally converge in \textit{interactive programming}, where systems must both reason about what code to generate and analyze execution results to guide fixes, optimizations, and capability expansions. This section explores how these reasoning capabilities have developed and synergized to build more sophisticated code intelligence systems.

\subsubsection{Reasoning for Code Generation}
\label{sec:r4c}
% logic: 
% CoT, instruction tuning, RLHF, RL (o1, deepseek r1) -> models gradually changing from base models to chat and then to today's "reasoning engine" -> Models usually think/plan first and then give answers, and code is no exception -> Many works have proved that reasoning can also significantly improve code generation, especially on more complex problems.
The integration of explicit reasoning has transformed code intelligence systems through advances in CoT~\cite{wei2023chainofthoughtpromptingelicitsreasoning}, instruction tuning~\cite{wei2022finetunedlanguagemodelszeroshot, muennighoff2024octopackinstructiontuningcode, luo2023wizardcoderempoweringcodelarge} and reinforcement learning~\cite{ouyang2022traininglanguagemodelsfollow, openai2024openaio1card, deepseekai2025deepseekr1incentivizingreasoningcapability}. Models have evolved from basic code completion tools~\cite{githubcopilot}, to applications with basic dialogue capabilities~\cite{openai_chatgpt}, and finally to sophisticated reasoning engines that combine planning, reasoning and critical thinking to arrive at solutions~\cite{openai2024openaio1card}, excelling at complex programming tasks.

Models adopt CoT reasoning as the core strategy, generating step-by-step thoughts before implementing code. Basic CoT improves code generation by articulating intermediate logic, while recent advancements adapt it to programming contexts, structuring reasoning around programmatic constructs (e.g., loops, conditionals) for correctness~\cite{li2023structuredchainofthoughtpromptingcode}, decomposing solutions into reusable modules for iterative refinement~\cite{huang2024codecottacklingcodesyntax}, and integrating problem decomposition for debugging~\cite{wen2024learningtaskdecompositionassist}. Models also generate natural language plans to guide implementation, ensuring alignment between intent and code logic~\cite{jiang2024selfplanningcodegenerationlarge, wang2024planningnaturallanguageimproves}. These strategies extend to resource-efficient scenarios, where lightweight models generate CoT steps through automated alignment frameworks~\cite{yang2024chainofthoughtneuralcodegeneration}, and to repository-level tasks, combining multi-step planning with static dependency analysis and code editing~\cite{bairi2023codeplanrepositorylevelcodingusing}. By integrating CoT with modular reasoning and context-aware planning, modern models achieve higher correctness and robustness in complex scenarios.

\subsubsection{Reasoning Over Code}
\label{sec:roc}
% DIFF:
% given input, ask outpu -> "reasoning over repo"
% given code, ask question ->  "reasoning on code"
While reasoning capabilities improve code generation, the ability to reason over code - understanding its behavior, predicting its execution, and analyzing its properties - remains a fundamental challenge in code intelligence. Unlike natural language, code's combination of rigid syntax with complex runtime behaviors demands comprehension of both static forms and dynamic execution, further complicated by external dependencies. Empirical studies show models can generate syntactically correct code while failing to grasp semantic meaning~\cite{zhu2024getempiricalstudycode}, highlighting the gap between surface manipulation and true understanding.

Various benchmarks assess different aspects of code reasoning: CRUXEval~\cite{gu2024cruxevalbenchmarkcodereasoning} focuses on predicting inputs from outputs and vice versa for simple Python functions, CodeMMLU~\cite{manh2024codemmlumultitaskbenchmarkassessing} assesses broader comprehension such as code analysis and defect detection, CodeQA~\cite{liu2021codeqaquestionansweringdataset} evaluates natural language reasoning about code, and RepoQA~\cite{liu2024repoqaevaluatinglongcontext} examines comprehension of repository-level code across multi-file contexts. CodeMind~\cite{liu2024codemindframeworkchallengelarge} introduces inductive reasoning tasks for evaluating execution understanding and specification adherence. To bridge the gap between surface-level code manipulation and deep understanding, recent approaches emphasize tracking and reasoning about program execution. NExT~\cite{ni2024nextteachinglargelanguage} teaches models to analyze runtime traces and generate explanatory rationales about program behavior, while SelfPiCo~\cite{xue2024selfpicoselfguidedpartialcode} enables reasoning about partial code through interactive execution simulation and state tracking. These approaches highlight understanding code requires more than static analysis - it demands the ability to mentally simulate program execution and track state changes over time. 
% While progress has been made through techniques like execution trace analysis and interactive simulation, bridging the gap between surface-level code manipulation and deep comprehension remains a fundamental challenge in advancing code intelligence systems.

% "reasoning to code + reason over code; interactively"
\subsubsection{Interactive Programming}
\label{sec:ip}
Recent researches enabled LLMs to autonomously evaluate and improve their outputs, with Self-Refine~\cite{madaan2023selfrefineiterativerefinementselffeedback} demonstrated how models can generate, critique, and optimize outputs. In code development, this mechanism gains unique advantages via the executable nature of code which provides immediate, objective feedback that triggers new reasoning cycles. Specifically, interactive programming forms a reasoning-driven optimization loop: models first reason to generate code for execution, then analyze execution results to understand errors or improvement directions, ultimately reasoning about better solutions. This embraces software development's iterative nature, advancing beyond traditional one-pass generation.

Early explorations in interactive program synthesis demonstrated feedback's potential\cite{le2017interactiveprogramsynthesis}, the emergence of LLMs catalyzed evolution to autonomous refinement: Self-Edit developed a fault-aware code editor leveraging execution results for iterative error correction~\cite{zhang2023selfeditfaultawarecodeeditor}, while InterCode established a comprehensive benchmark environment and standardized interactive coding as a reinforcement learning problem~\cite{yang2023intercodestandardizingbenchmarkinginteractive}.
Recent advances have further refined this paradigm: CodeChain introduced self-revision mechanism that modularizes code generation and systematically improves solutions through targeted refinement chains~\cite{le2024codechainmodularcodegeneration}, LeTI demonstrated improvement through natural language feedback~\cite{wang2024letilearninggeneratetextual}, and OpenCodeInterpreter unified generation, execution, and refinement in one framework~\cite{zheng2025opencodeinterpreterintegratingcodegeneration}. Systematic analysis reveals these methods' effectiveness heavily depends on models' ability to reason about program behavior and execution feedback~\cite{zheng2024makeslargelanguagemodels}. This evolution has laid crucial groundwork for code agents capable of handling complex programming tasks.

\subsection{Code Agents with Complex Reasoning}
\label{sec:agent}
% has CoT, reasoning on code, interactive...all kind of techniques....the combinition of all aforementioned techniques. (but with additional features interactive with env)

The convergence of code reasoning paradigms -- planning and decomposition, context-aware understanding, and interactive programming -- has enabled the evolution of code intelligence systems into autonomous code agents~\cite{devinAI, cursorAI, openhands2024}. These agents handle complex development tasks by decomposing tasks and formulating execution plans, translating abstract solutions into concrete environmental actions through predefined tools (e.g., IDE operations, terminal commands), and continuously monitoring execution states while gathering environmental feedback to reach goals. Unlike static code generators, these agents treat development as a dynamic decision cycle by interacting with the environment, with reasoning applied throughout—from understanding requirements and taking appropriate actions to evaluating outcomes.

SWE-bench established a comprehensive evaluation framework based on real GitHub issues~\cite{jimenez2024swebenchlanguagemodelsresolve}, later expanded with SWE-bench Multimodal~\cite{yang2024swebenchmultimodalaisystems} incorporating visual software tasks and SWE-bench Verified~\cite{chowdhury2024swebenchverified} enhancing evaluation reliability through rigorous test case validation. These evaluations revealed persistent challenges in code intelligence: effective reasoning about program structure and behavior, safe and effective codebase navigation and modification, and maintaining coherent long-term planning across development iterations.

Modern code agents share a common foundation in environment interaction, while each contributing unique implementation focuses. CodeAct~\cite{wang2024executablecodeactionselicit} pioneered executable agent behaviors through Python interpreter, enabling dynamic debugging workflows, and OpenHands~\cite{openhands2024} extended it by providing a flexible agent infrastructure supporting customizable tool chains. SWE-agent~\cite{yang2024sweagentagentcomputerinterfacesenable} focused on optimizing repository navigation through Agent-Computer Interface,  CodeAgent~\cite{zhang2024codeagentenhancingcodegeneration} combined tool specialization with strategic frameworks, coordinating multiple repository-level operations and AutoCodeRover~\cite{zhang2024autocoderoverautonomousprogramimprovement} introduced spectrum-based fault localization to guide context retrieval.

Recent advances have explored two contrasting directions: multi-agent systems and agent-free approaches. HyperAgent~\cite{phan2024hyperagentgeneralistsoftwareengineering} coordinates specialized agents for planning, navigation, editing, and execution, demonstrating how different reasoning capabilities can be hierarchically orchestrated. In contrast, Agentless~\cite{xia2024agentlessdemystifyingllmbasedsoftware} achieves effectiveness through simplification - employing a focused two-phase process for fault localization and repair without complex agent architectures. Empirical evaluations show that, compared to humans, these approaches reduce code redundancy, with effective task decomposition being key to success, ~\cite{chen2024evaluatingsoftwaredevelopmentagents}, though matching human-level performance remains challenging.

\section{Challenges and Future Directions}
\label{sec:fut}
\vspace{-2mm}

\subsection{Code-enhanced Reasoning}

\noindent\textbf{Lack of Interpretability and Debuggability.} A key challenge in code-enhanced reasoning is the reliance on the code generation capabilities of LLMs~\citep{kadavath2022language, kabra2023program}. However, LLM-generated code often does not faithfully reflect the model's true chain of thought~\citep{li2023chain}, nor can these models reliably assess their own confidence~\citep{kabra2023program}. Manual inspections of the generated code are time-consuming and prone to oversight~\citep{li2023chain, tian2023test}, underscoring the need for systematic error detection and robust error-handling strategies within the code itself~\citep{li2023chain, ni2024next}.
Mechanisms that empower LLMs to self-reflect and debug their generated code would be highly beneficial~\citep{chen2024steering}. Potential approaches include tree-based generation~\citep{yao2023tree}, reasoning-oriented self-reflection~\citep{shinn2023reflexion}, and reinforcement learning methodologies~\citep{le2022coderl}. Another promising avenue is the application of formal verification techniques~\citep{kang2025explainable}, which can validate the correctness of the generated code and ensure alignment between the code logic and intended reasoning steps.

% A primary challenge in code-enhanced reasoning is the relianve on the code generation capabilities of LLMs~\citep{kadavath2022language, kabra2023program}. However, LLM generated code sometimes doesn't faithfully reflect the model's reasoning~\cite{li2023chain} and fail to assess their own confidence---effectively ``not knowing what they do not know''~\cite{kabra2023program}. Manual inspection also remains time-consuming and error-prone~\citep{li2023chain, tian2023test}, highlighting a need for systematic error detection, robust error-handling within code, and representations less vulnerable to small mistakes~\citep{li2023chain, ni2024next}.
% Future work should implement mechanisms for LLMs to self-reflect on and debug generated code~\cite{chen2024steering}. This could involve techniques such as tree-based generation~\cite{yao2023tree}, reasoning-oriented self-reflection~\cite{shinn2023reflexion} and reinforcement learning~\cite{le2022coderl}. In addition, research into formal verification methods to validate the correctness of the generated code and ensure consistency between the code and the intended reasoning steps would be valuable~\cite{kang2025explainable}.

\noindent\textbf{Blended Code-and-Language Reasoning.} Although code excels at numeric and algorithmic tasks, it frequently struggles with less structured or more subjective tasks (e.g., commonsense reasoning, semantic analysis) where purely executable representations are inadequate~\citep{li2023chain, weir2024learning, liu2024interactive}. A crucial challenge is deciding how to split reasoning processes between structured code (for precise computation) and free-form text (for broader contextual and interpretive functions)~\citep{suzgun2022challenging, liu2024interactive, xiong2024inc}.
Frameworks such as ``LMulator'' and ``pseudocode execution'' demonstrate the potential of interleaving code generation with textual reasoning~\citep{li2023chain, weir2024learning}, allowing symbolic computation to be complemented by natural language interpretation. Moving forward, designing hybrid architectures that seamlessly integrate code and language modalities will be essential for improving performance on a wide range of tasks, particularly those requiring nuanced judgment alongside algorithmic precision.

% Code-based methods excel at numeric and algorithmic tasks but struggle with ``fuzzy'' domains (e.g., commonsense or subjective evaluations) where purely executable representations are inadequate~\citep{li2023chain, weir2024learning, liu2024interactive}. A critical open challenge is to partition reasoning processes into structured code (for precise computation) and free-form text (for contextual interpretation)\citep{suzgun2022challenging, liu2024interactive, xiong2024inc}. Emerging frameworks such as ``LMulator'' and ``pseudocode execution'' interleave code generation and textual reasoning to address this gap\citep{li2023chain, weir2024learning}, suggesting a promising direction for hybrid architectures.

\noindent\textbf{Optimizing Code Data and Representations} Determining the optimal level of code complexity for enhancing reasoning remains an open problem. Overly intricate code can be difficult for LLMs to learn effectively, while overly simplistic code may fail to capture essential reasoning steps~\citep{bi2023when}.
A systematic analysis of the relationship between code complexity and reasoning performance is needed. Metrics such as cyclomatic complexity and code length can help quantify code difficulty and guide the selection of complexity levels that maximize learning efficiency. Additionally, adaptive curricula that gradually increase code complexity may enable LLMs to progressively acquire more sophisticated reasoning capabilities while minimizing the risk of overwhelming the model.

% Determining the optimal level of code complexity for improving reasoning abilities is a challenge. Overly complex code can be too difficult for LLMs to learn, while overly simplistic code may not capture the full reasoning process~\cite{bi2023when}. Future research could Conduct a systematic analysis of the relationship between code complexity and reasoning performance. This could involve using metrics such as cyclomatic complexity and code length to quantify code complexity and identify the optimal range for different types of tasks.

\noindent\textbf{Lack of Scalability and Generalization.} Many current code-enhanced reasoning methods rely on task-specific fine-tuning, which can hinder generalization to novel tasks or domains~\citep{yu2023metamath, mitra2024orca, wang2023mathcoder}. Moreover, data scalability often remains limited to narrow domains (e.g., mathematical calculation, code manipulation)~\citep{guo2024deepseek, hui2024qwen25codertechnicalreport, lozhkov2024starcoder2stackv2, laurenccon2022bigscience, wen2024unlocking}, restricting the applicability of these models in real-world scenarios.
Improving zero- and few-shot learning capabilities will be crucial for broadening the scope of code-enhanced reasoning~\citep{chen2022program}. Innovative data augmentation techniques, such as generating synthetic data or leveraging unsupervised learning on unlabeled corpora, can further enrich model training~\citep{phan2023training, lightman2023let}. Finally, cross-domain training strategies~\citep{li2023self} that integrate knowledge from multiple sources hold promise for more robust, generalized reasoning across diverse tasks and domains.

% Many current code-enhanced reasoning methods involve task-specific fine-tuning, which often results in poor generalization to new tasks or domains~\cite{yu2023metamath, mitra2024orca, wang2023mathcoder}. In addition, many LLMs are constrained by data scalability, significantly limited to narrow domains such as Math calculation and Code~\citep{guo2024deepseek, hui2024qwen25codertechnicalreport, lozhkov2024starcoder2stackv2, laurenccon2022bigscience}, which limits their applicability to more complex, real-world scenarios~\cite{wen2024unlocking}. Consequently, building a scalable dataset and generalized reasoning capability of LLMs across different domains are essential.
% Future research may focus on improving the zero/few-shot learning capabilities of code-enhanced reasoning systems. This involves developing methods that allow the models to quickly adapt to new tasks with only a few examples~\cite{chen2022program}. Also, developing innovative data augmentation strategies are also worth explore. This could involve generating synthetic data or leveraging unsupervised learning techniques to extract knowledge from unlabeled data~\cite{phan2023training, lightman2023let}. Also, Investigate cross-domain training approaches that enable models to learn from multiple domains simultaneously, improving their ability to generalize to new, unseen domains~\cite{li2023self}.

\noindent\textbf{Difficulty with Complex or Abstract Tasks} While code-based approaches excel in structured problem-solving, they often falter on tasks requiring commonsense, semantic interpretation, or complex algebraic reasoning. In some instances---such as evaluating the humor in a name edit---code-based reasoning may even introduce unnecessary complexity or degrade performance~\citep{li2023chain}.
Next-generation models should be designed to be more context-aware, capable of determining when code is beneficial and when alternative strategies would be more appropriate~\citep{chen2024steering}. Achieving this requires adaptive, multimodal architectures that selectively combine code execution with natural language processing and other reasoning paradigms, ensuring that different task types receive the most effective mode of reasoning support.

\noindent\textbf{Lack of High-Quality Datasets.} Many open-
source code LLMs still rely on training data scraped from GitHub, which can suffer from redundancy, poor quality, and overly short snippets~\citep{deepseekai2024deepseekcoderv2breakingbarrierclosedsource, hui2024qwen25codertechnicalreport, lozhkov2024starcoder2stackv2}. Consequently, building cleaner and more diverse datasets is essential for advancing tasks such as code generation and editing. High-quality dataset curation not only improves model performance but also benefits the broader community seeking robust benchmarks and reproducible experimental settings

\noindent\textbf{Tool Usage Based on Code Format} Currently, LLMs or agents typically use APIs or simple code to invoke tools~\citep{shen2023hugginggpt, qin2023toolllm}. However, in complex working conditions, the construction of a sophisticated and complete tool usage chain remains an unsolved challenge. Code, as a universal format, has a unique advantage in this aspect. The key question is how to design a standardized format that enables LLMs or agents to invoke available tools on a computer through automated code generation and execution. This approach enhances the capabilities of LLMs or agents, allowing them to tackle more complex tasks effectively.

% While code-enhanced reasoning excels in structured problem-solving, it often struggles with tasks requiring commonsense reasoning, semantic analysis, or complex algebraic reasoning, Some tasks, like determining if a name edit is humorous, may not benefit or even harm from code at all~\cite{li2023chain}. Future models should be designed to be more context-aware, capable of discerning when code is beneficial and when it is not, and adapting their approach accordingly~\cite{chen2024steering}

\subsection{Reasoning-enhanced Code Intelligence}

\noindent\textbf{Large-Scale Code Understanding}  
Large-scale code understanding has seen significant progress with the expansion of context windows, enabling models to process even over 1 million tokens~\cite{chen2023extendingcontextwindowlarge, guo2023longcoderlongrangepretrainedlanguage}. However, increasing context length does not always lead to better comprehension, as models struggle to focus on critical information when relevant code snippets are dispersed across a repository~\cite{li2024looglelongcontextlanguagemodels}. Retrieval-Augmented Generation (RAG) has been introduced to mitigate this issue by retrieving relevant segments, but it is not without limitations: key information may be missed, and retrieval strategies may not always align with complex code structures~\cite{wu2024pandorasboxaladdinslamp, jin2024longcontextllmsmeetrag, yu2024defenserageralongcontext}. Striking a balance between context expansion, retrieval augmentation, and precise code parsing is essential to building product-grade code intelligence systems capable of both global comprehension and accurate localization, making them effective for complex repository-level tasks.

\noindent\textbf{Long-Form Code Generation}
Recent advances in LLMs for code generation have primarily focused on handling longer input contexts rather than generating longer, structured code outputs~\cite{wu2025longgenbenchbenchmarkinglongformgeneration}. In other words, current training optimizes long-context understanding, but does not necessarily improve the coherence and quality of long-form code generation. Several challenges arise in long-form generation: first, it is difficult to evaluate, as most existing benchmarks assess the correctness of individual functions, while assessing multi-file, multi-module code remains an open problem. Second, long-form code generation is prone to errors—when the output scale increases, the accumulation of small mistakes can render the entire project non-functional or logically inconsistent. Moreover, correctness and executability are difficult to ensure, as large-scale software development involves rigorous compilation, testing, and debugging processes, which generated code may not adhere to. Future research should focus on improving training strategies for long-form generation, developing better evaluation metrics for multi-file coherence, and ensuring correctness and executability in large-scale code generation.

\noindent\textbf{Exploring the Applicability of Reasoning Models in Code Agents}
Despite significant breakthroughs in mathematical reasoning and code generation, reasoning models such as O1 and R1~\cite{openai2024openaio1card, deepseekai2025deepseekr1incentivizingreasoningcapability, openai2025o3mini} have shown limited improvements in agent-based tasks. One possible explanation is that existing agent frameworks were optimized for earlier non-reasoning models, which prevents newer models from fully leveraging their reasoning capabilities. Alternatively, reasoning-enhanced models may not inherently excel in agent-based tasks, meaning their strengths in mathematical and code reasoning do not necessarily translate into superior agent execution. If the latter is true, adapting agent architectures alone may not be sufficient, and a more fundamental investigation into the role of reasoning models in agents is needed. Future research should explore new agent frameworks, better utilization of reasoning capabilities, and empirical validation of reasoning-enhanced models in real-world programming agent scenarios to determine whether new paradigms are required or if models themselves need refinement to be more effective in agent environments.

\noindent\textbf{Balancing Autonomy and Control in Code Agents}  
As agents become more capable, the balance between autonomy and control emerges as a crucial challenge. Allowing agents more freedom to explore solutions independently may yield novel and highly efficient results, while enforcing strict control mechanisms ensures predictability and reliability. Finding the right balance between these approaches is essential for practical deployment. Additionally, safety concerns grow with increased agent autonomy, particularly in scenarios involving direct code execution~\cite{guo2024redcoderiskycodeexecution}. Intelligent safeguards are needed to prevent security vulnerabilities, unintended execution of high-risk operations, and harmful self-modifications. Future research should investigate frameworks that enable agents to operate within safe execution environments while maximizing their ability to autonomously optimize and improve code generation.  

\noindent\textbf{Multimodal Code Intelligence} The evolution of programming from purely text-based workflows to multimodal interactions is reshaping the development landscape, particularly in UI/UX and frontend engineering~\cite{yun2024web2codelargescalewebpagetocodedataset}. Traditional code models primarily rely on textual inputs, but future systems will require capabilities to process visual elements, bridging the gap between design and implementation. Advancements in aesthetic-aware LLMs~\cite{abe2024assessingaestheticevaluationcapabilities}, vision-based coding agents~\cite{zheng2023seeact}, and interface manipulation technologies~\cite{anthropic2024computeruse} offer exciting possibilities. Future research should focus on training models that can generate code from visual specifications, interact with IDEs through graphical interfaces, and develop datasets that capture the intricate relationships between design components and their code representations, paving the way for more intuitive and efficient development workflows.

\noindent\textbf{Reinforcement Learning for Code Models}  
Reinforcement learning (RL) presents a promising avenue for enhancing reasoning in code models. Unlike other domains, code execution provides immediate and objective feedback, making it well-suited for RL-based optimization. One potential approach involves training models to predict input-output behavior for given code and test cases, using CoT reasoning expressed in natural language to guide the learning process~\cite{deepseekai2025deepseekr1incentivizingreasoningcapability}. Another key direction is exploring RL in agent-based environments, where agents can iteratively refine their strategies for code search, debugging, and refactoring through trial and error. Incorporating RL into code intelligence systems may significantly enhance their reasoning depth, problem-solving efficiency, and overall robustness.  

\noindent\textbf{Innovation and Refinement of Evaluations}
As code intelligence models continuously master existing benchmarks~\cite{xia2024leaderboardrankingcoding}, the development of new evaluation frameworks remains a perpetual necessity~\cite{mcintosh2024inadequacieslargelanguagemodel}. Future research must create more sophisticated benchmarks that better reflect real-world challenges while resisting data contamination~\cite{riddell2024quantifyingcontaminationevaluatingcode}. These frameworks should also extend beyond mere functional correctness to assess broader software development aspects, e.g., code quality, maintainability, and design aesthetics~\cite{simões2024evaluatingsourcecodequality, borg2024doescodevelopmentaiassistants}.

\section{Conclusion}

The synergy between code and reasoning has driven significant advancements in AI, with code enhancing logical reasoning and reasoning improving code intelligence. This survey explored how executable programs and structured code paths refine AI reasoning while highlighting how reasoning abilities enables advanced code generation, comprehension, and debugging.  Despite progress, challenges such as ambiguity, scalability, and consistency remain. Future research must deepen the integration of reasoning and programming to build more robust, interpretable, and adaptive AI systems. As these fields converge, AI’s ability to think and code will continue to evolve, reshaping intelligent automation.

\section{Limitations}

Our survey spans a wide range of approaches, from single-execution code-based reasoning (\S\ref{sec:c4r}) to advanced autonomous code agents (\S\ref{sec:agent}), which compels us to keep certain implementation details and domain-specific nuances only briefly described. The decision to focus on recent arXiv categories and a confined publication window excludes older or less mainstream work that could offer alternative perspectives or historical context. Coverage of benchmarks mentioned in \S\ref{sec:roc} and \S\ref{sec:agent}—CRUXEval, CodeMMLU, RepoQA, and SWE-bench—remains incomplete with respect to real-world repository-scale tasks or specialized areas such as concurrency analysis and security verification. The challenges identified in \S4 reflect ongoing research gaps rather than definitive conclusions, and future developments in datasets, model architectures, and evaluation protocols may prompt revisions or expansions of this survey.

\bibliography{custom}

\appendix

% \section{Related Survey}
% \label{apsec:rs}

% \section{Further Discussion}
% \label{apsec:fd}

\section{Technical Introduction for Important Methods}
\label{apsec:method}

\subsection{Code-enhanced Reasoning}

In this section, we provide additional technical insights into how code-generation 
strategies serve as a scaffolding mechanism for complex reasoning. By interleaving textual 
explanations with executable or pseudo-executable code, these methods leverage the language 
model's ability to decompose tasks while offloading precise computations to interpreters 
or simulators. Below, we outline four representative approaches.

\paragraph{Program-Aided Language Models (PaL)}
PaL~\cite{gao2023pal} interleaves natural language reasoning and programmatic statements by prompting large 
language models to emit both text (e.g., comments) and code (e.g., Python snippets). 
Any arithmetic or logical operations are delegated to a code interpreter, allowing the 
model to focus on higher-level step-by-step reasoning rather than raw calculation. 
This reduces errors in multi-step tasks, as correctness is grounded in the verified 
outputs from executing the code.

\paragraph{Program of Thoughts (PoT)}
PoT~\cite{chen2022program} frames the solution process as the generation of a "program of thoughts," 
where each sub-step is encoded in semantically meaningful variables and partial code. 
Once generated, the code is executed externally to reliably produce numerical 
results. By breaking down complex computations into a series of small, interpretable 
code snippets, PoT enables more transparent and robust multi-step reasoning.

\paragraph{MathCoder}
MathCoder~\cite{lu2024mathcoder2} provides a dynamic interplay between reasoning and real-time code execution. 
The model switches between producing language-based rationales and code blocks, 
executing each snippet as it is generated. The output of each block is then folded 
back into the ongoing chain of thought, resulting in an iterative loop of code-based 
calculation and textual reasoning that can tackle intricate math problems more reliably.

\paragraph{Chain of Code (CoC)}
CoC~\cite{li2023chain} mixes semantic reasoning and code-like structures, but allows certain segments 
of generated code to be ``emulated'' by the language model itself if they are not 
executable in a standard interpreter. Whenever actual code execution is possible, 
it is performed directly (e.g., for arithmetic). Otherwise, the language model 
simulates the code’s effect, maintaining a consistent state. This hybrid approach 
combines symbolic execution with language-driven inference for tasks that blend 
logical, numerical, and semantic reasoning.

\subsection{Training with Code}

In this section, we illustrate five noteworthy methods that harness code-generation to bolster reasoning capacity.These approaches use code data for training to structure the thinking process, verify intermediate steps, and produce more precise final answers.

\paragraph{MARIO}
MARIO~\cite{liao2024mario} addresses the challenge of enhancing mathematical reasoning in LLMs by introducing an enriched math dataset derived from GSM8K and MATH, refined through GPT-4 annotations, human review, and self-training. Central to its approach is the utilization of a Python code interpreter, enabling models to perform exact calculations and systematic error checks. MARIO also proposes a replicable fine-tuning protocol that substantially improves performance on GSM8K and MATH. By making both the source code and trained models publicly available, MARIO contributes an open, community-driven platform for advancing code-based mathematical reasoning.

\paragraph{POET}
POET~\cite{pi2022reasoning} boosts a model’s reasoning capacity by pretraining it on programs and their execution results, effectively importing a ``program executor's'' knowledge into the language modeling process. Instantiated as POET-Math, POET-Logic, and POET-SQL, it covers numerical, logical, and multi-hop reasoning tasks. Through data-driven alignment of natural language and code, POET significantly strengthens a model’s ability to conduct step-by-step inferences and validate conclusions.

\paragraph{CodePMP}
CodePMP~\cite{yu2024codepmp} proposes a scalable preference model pretraining pipeline that leverages large corpora of synthesized code-preference pairs. By training reward models on these code-centric preferences, CodePMP eases the scarcity of human-labeled data and refines LLMs’ reasoning via reinforcement learning from human feedback. Experiments on mathematical reasoning (GSM8K, MATH) and logical reasoning (ReClor, LogiQA2.0) show notable improvements, highlighting the value of code-based preference modeling for multi-step inference tasks.

\paragraph{SIAM}
SIAM~\cite{yu2024siam} targets code-centric mathematical problem-solving by tapping into large-scale, expert-written math question-answer pairs and enforcing rigorous quality checks through a code-based critic model. Beyond merely augmenting GSM8K-like data, SIAM refines alignment via self-generated instruction and preference data, preventing narrow overfitting to specific question types. The approach consistently boosts performance across both in-domain and out-of-domain math benchmarks, in multiple languages, showcasing robust generalization in code-enhanced reasoning.

\paragraph{CODEPLAN}
CODEPLAN~\cite{bairi2023codeplanrepositorylevelcodingusing} tackles multi-step reasoning bottlenecks by introducing ``code-form plans,'' or structured pseudocode, as intermediate representations. This framework enables LLMs to outline and execute high-level reasoning flows, capturing control structures and semantic details often missing in plain text. Trained on a large-scale dataset of paired plan-response examples, CODEPLAN delivers substantial gains across diverse tasks including mathematical, symbolic, multi-hop QA, and decision-making scenarios. Its data-efficient and lightweight design underscores the advantage of code-form reasoning for complex problem-solving.

\subsection{Reasoning-enhanced Code Intelligence}

\noindent
This section examines prominent approaches that integrate reasoning capabilities into code generation. These methods span a spectrum of techniques including planning and task decomposition, self-improvement loops, interactive refinement processes, and agent-based frameworks. By incorporating sophisticated reasoning mechanisms, these approaches aim to enhance the quality, reliability, and maintainability of generated code while addressing complex programming challenges across different contexts and scales.

\paragraph{Self-Planning}
Self-Planning~\cite{jiang2024selfplanningcodegenerationlarge} decomposes the generation process into two distinct phases. In the planning phase, the model generates a high-level plan from the task’s natural language intent using a few exemplars, and in the subsequent implementation phase, this plan guides the step-by-step synthesis of code. This division facilitates improved handling of complex code generation tasks by breaking down intricate requirements into manageable sub-tasks.

\paragraph{SCoTs}
SCoTs~\cite{li2023structuredchainofthoughtpromptingcode} refines traditional chain-of-thought methods by explicitly incorporating programming constructs—such as sequences, branches, loops, and input-output structures—into the intermediate reasoning. This structured approach directly aligns the model’s generated thought processes with the formal structure of code, leading to more robust, readable, and accurate code synthesis.

\paragraph{CodeCoT}
CodeCoT~\cite{huang2024codecottacklingcodesyntax} integrates chain-of-thought reasoning with a self-examination loop to target code syntax errors. After initially generating code via intermediate reasoning, the model produces test cases to validate syntax through local execution. Feedback from this self-testing phase is then used to iteratively refine the code, ensuring that the final output adheres to both logical consistency and strict syntactic requirements.

\paragraph{CodePlan}
CodePlan~\cite{bairi2023codeplanrepositorylevelcodingusing}formulates repository-level coding tasks as a planning problem by synthesizing a multi-step chain of edits that span multiple inter-dependent files. By leveraging incremental dependency analysis, change impact evaluation, and adaptive planning strategies, the framework orchestrates coordinated modifications across large codebases, thus automating complex repository-level transformations with higher accuracy and consistency.

\paragraph{COTTON}
COTTON~\cite{yang2024chainofthoughtneuralcodegeneration} enables lightweight language models (with fewer than 10 billion parameters) to benefit from high-quality chain-of-thought reasoning. By decoupling the generation of intermediate reasoning traces from the final code synthesis and leveraging externally generated CoTs, COTTON allows resource-efficient models to achieve performance gains comparable to those of much larger models.

\paragraph{PlanSearch}
PlanSearch~\cite{wang2024planningnaturallanguageimproves} incorporates explicit natural language planning into the code generation process. By prompting models to articulate detailed, coherent plans before commencing code synthesis, this method improves the search and selection of relevant code snippets, thus reducing errors and enhancing the overall quality of generated code in complex programming scenarios.

\paragraph{NExT}
NExT~\cite{ni2024nextteachinglargelanguage} introduces a framework that trains large language models to inspect execution traces—capturing variable states and control flows during runtime—and integrates these observations into chain-of-thought rationales. By self-training on synthetic execution-aware data, the method equips models with a semantic understanding of dynamic code behavior, which is then leveraged for improved program repair and debugging performance.

\paragraph{SelfPiCo}
SelfPiCo~\cite{xue2024selfpicoselfguidedpartialcode} leverages an interactive loop to convert non-executable code fragments into runnable snippets. It integrates few-shot in-context learning with chain-of-thought reasoning to predict appropriate dummy values for undefined elements and refines these predictions based on execution feedback. The framework is built around key components—including an interactive value predictor and a complementary type predictor—that work together to iteratively adjust and complete partial code segments, thereby transforming incomplete code into an executable form without altering existing code structure.

\paragraph{Self-Refine}
Self-Refine~\cite{madaan2023selfrefineiterativerefinementselffeedback} introduces an iterative self-feedback mechanism in which the same large language model first generates an initial output and then critiques and refines it through repeated feedback cycles. By interleaving a feedback phase that evaluates various aspects of the output with a subsequent refinement phase that corrects any identified shortcomings, the approach systematically enhances output quality. The method avoids the need for extra training data by leveraging few-shot prompting and untangling reasoning from correction, thereby improving performance across diverse tasks.

\paragraph{Self-Debugging}
Self-Debugging~\cite{chen2023teachinglargelanguagemodels} equips models with the ability to autonomously detect and repair errors in generated code. The method begins with an initial code generation step, followed by code execution that reveals runtime issues. The model then generates natural language explanations of the detected errors and revises its code accordingly. This self-debugging process, guided by few-shot demonstrations, effectively simulates a human debugging session and leads to more robust and accurate code synthesis.

\paragraph{Self-Collaboration}
Self-Collaboration~\cite{dong2024selfcollaborationcodegenerationchatgpt} employs a simulated internal dialogue where the model engages in self-interaction to revise and consolidate its code output. By using chain-of-thought prompting, ChatGPT generates multiple reasoning iterations that simulate collaborative discussion, enabling it to reconcile different coding strategies. This self-collaborative approach improves the precision and resilience of generated code through iterative internal debate and refinement.

\paragraph{Self-Edit}
Self-Edit~\cite{zhang2023selfeditfaultawarecodeeditor} incorporates a dedicated fault detection phase into the code generation process. After producing an initial draft, the system analyzes the code for syntactic and semantic errors, annotating potential faults. The model then utilizes this fault-aware feedback to perform targeted edits that correct mistakes and optimize functionality. This iterative loop of analysis and refinement results in higher-quality code that is both more efficient and bug-resistant.

\paragraph{LeTI}
LeTI~\cite{wang2024letilearninggeneratetextual} redefines code generation as an interactive, dialogue-driven process. By capturing multi-turn textual interactions, the framework aggregates diverse reasoning cues and iteratively refines code outputs. The model uses conversational context and chain-of-thought reasoning to integrate these insights, which enhances both the interpretability and accuracy of the final code. This process promotes a more holistic synthesis of programming solutions based on natural language reasoning.

\paragraph{InterCode}
InterCode~\cite{yang2023intercodestandardizingbenchmarkinginteractive} proposes a standardized framework that embeds real-time execution feedback into the coding process. By systematically incorporating dynamic execution results into iterative refinement cycles, the approach establishes benchmarks for interactive coding performance. The integration of execution trace analysis ensures that the feedback loop directly informs code corrections, thereby raising the reliability and robustness of generated code in practical software development contexts.

\paragraph{CodeChain}
CodeChain~\cite{le2024codechainmodularcodegeneration} adopts an iterative self-revision strategy to decompose complex programming tasks into modular sub-tasks. Initially, the model generates modularized code using chain-of-thought prompting. It then extracts and clusters sub-modules from the generated code, selecting representative components that are reintroduced into subsequent prompts. This cycle enables the model to refine its solutions through reuse of verified sub-modules, enhancing both the modularity and correctness of the final output.

\paragraph{AgentCoder}
AgentCoder~\cite{huang2024agentcodermultiagentbasedcodegeneration} formulates code generation as a collaborative multi-agent process wherein different agents specialize in distinct roles. One agent generates an initial code draft, another evaluates its correctness through testing, and a third optimizes performance based on iterative feedback. The interplay among these agents, facilitated by competition and collaboration, continuously refines the generated code until an optimal solution is reached.

\paragraph{OpenCodeInterpreter}
OpenCodeInterpreter~\cite{zheng2025opencodeinterpreterintegratingcodegeneration} bridges the gap between static code synthesis and dynamic validation by integrating code generation with immediate execution feedback. The method prompts the language model to produce code, which is then directly executed to obtain runtime results. These outcomes inform iterative refinement cycles, allowing the model to adjust its generated solutions based on real-time execution data, ultimately leading to more reliable and performant code.

\paragraph{CodeAgent}
CodeAgent~\cite{zhang2024codeagentenhancingcodegeneration} decomposes repo-level code synthesis into a series of coordinated tool invocations. Its technical framework integrates external programming tools—such as information retrieval, code symbol navigation, format checking, and code interpretation—with multiple agent strategies (e.g., ReAct, Tool-Planning, OpenAIFunc, and rule-based usage). This modular design allows the LLM to dynamically leverage these tools, iteratively refine its outputs, and generate cohesive code for complex codebases.

\paragraph{CodeAct}
CodeAct~\cite{wang2024executablecodeactionselicit} reformulates LLM agent behavior by consolidating actions as executable Python code. By harnessing Python’s native control and data flow constructs, the method enables multi-turn interactions where code execution feedback—ranging from success signals to error tracebacks—is used to iteratively revise and improve subsequent actions. This technical shift from rigid JSON/text formats to dynamic code actions streamlines tool composition and self-debugging.

\paragraph{AutoCodeRover}
AutoCodeRover~\cite{zhang2024autocoderoverautonomousprogramimprovement} presents an autonomous loop for program improvement, where the LLM continually refines its generated code. The system employs runtime feedback and error analysis to detect deficiencies, triggering self-debugging routines and automated optimizations. By iteratively re-running the code and integrating improvements, AutoCodeRover progressively enhances program correctness and efficiency within a closed-loop refinement process.

\paragraph{SWE-agent}
SWE-agent~\cite{yang2024sweagentagentcomputerinterfacesenable} constructs an interactive interface that mimics developer workflows for software engineering tasks. Its technical approach centers on integrating LLM-driven tool invocation with environments that supply real-time code dependency analysis, automated testing, and validation. This design empowers the agent to traverse complex code ecosystems, where iterative tool-guided feedback enables continuous adjustments and reliable code synthesis.

\paragraph{Agentless}
Agentless~\cite{xia2024agentlessdemystifyingllmbasedsoftware} challenges the necessity of explicit agent orchestration by embedding tool interaction directly into the LLM’s reasoning process. Using an agent-free paradigm, it leverages chain-of-thought reasoning alongside direct tool calls, reducing structural overhead while still ensuring context-aware code generation and debugging. This minimalist design streamlines the coding process by allowing the LLM to self-manage multi-turn interactions without dedicated intermediary agent modules.

\paragraph{OpenHands}
OpenHands~\cite{openhands2024} offers a modular, open platform that empowers AI software developers by integrating a diverse suite of development tools. Its technical architecture provides a unified interface for tool selection, code generation, and interactive debugging, enabling LLMs to perform repo-level tasks and collaborative scenarios. By fusing native code execution with flexible action orchestration, OpenHands facilitates seamless transitions between varied software engineering challenges.

\paragraph{HyperAgent}
HyperAgent~\cite{phan2024hyperagentgeneralistsoftwareengineering} scales LLM-based software engineering by adopting hierarchical task decomposition and parallel tool integration. Its framework orchestrates multiple specialized sub-agents coordinated via dynamic feedback loops, enabling the simultaneous handling of extensive coding tasks. By leveraging multi-agent collaboration and real-time code refinement, HyperAgent achieves robust, scalable performance across complex programming environments.

\section{Introduction of Important Benchmarks}
\label{apsec:bench}

\subsection{Code-enhanced Reasoning}

The emergence of code-enhanced mathematical reasoning has motivated the development of specialized datasets to evaluate models' reasoning capabilities. While the main paper discusses the methodological advances, this section provides detailed characterizations of three representative datasets that have significantly shaped this research direction. These datasets are particularly noteworthy for their distinct approaches to assessing reasoning.

\paragraph{GSM8K} GSM8K~\cite{cobbe2021gsm8k} contains 8.5K grade school math word problems requiring 2-8 steps of reasoning to solve. The problems are designed to have high linguistic diversity while relying on elementary mathematical concepts. The dataset emphasizes multi-step deductive reasoning rather than complex mathematical knowledge, with natural language solutions that explicitly demonstrate the step-by-step reasoning process.

\paragraph{MATH} MATH~\cite{hendrycks2021math} comprises 12,500 competition mathematics problems drawn from various sources including AMC 10, AMC 12, and AIME. Unlike GSM8K which focuses on elementary reasoning, MATH problems require more sophisticated mathematical problem-solving heuristics and domain knowledge. Each problem in MATH comes with a detailed step-by-step solution that demonstrates both mathematical reasoning and domain-specific problem-solving strategies.

\paragraph{SVAMP} SVAMP~\cite{patel2021svamp} is a challenge set of 1,000 problems designed to test the robustness of reasoning capabilities in math word problem solvers. While maintaining similar mathematical complexity to existing datasets, SVAMP introduces systematic variations along three key dimensions: question sensitivity (testing if models truly understand the question), reasoning ability (testing if models can adapt to subtle changes requiring different reasoning paths), and structural invariance (testing if models maintain consistent reasoning across superficial changes).

\subsection{Training with Code}

\noindent
This section provides concise technical overviews of key benchmarks that have significantly guided code-based reasoning research. These datasets distinguish themselves through various approaches—ranging from multi-hop textual analysis to environment-based decision-making—all designed to rigorously evaluate a model’s reasoning capabilities.

\paragraph{OCW}
OCW~\cite{lewkowycz2022solving} is designed to test a model’s ability to reason through open-ended questions that often require code-based logic or structured problem-solving. It presents a mix of prompts that may include mathematics, algorithmic puzzles, or short coding snippets, pushing models to generate reasoned solutions rather than superficial answers. As such, it emphasizes step-by-step thinking and logical correctness.

\paragraph{HotpotQA}
HotpotQA~\cite{yang2018hotpotqa} is a multi-hop question-answering dataset that requires a model to connect information across multiple documents or sentences to arrive at a correct response. Its emphasis on evidence-based reasoning makes it a strong benchmark for evaluating how well models can chain together relevant facts logically. While not code-focused, it indirectly supports code-enhanced approaches by encouraging structured, stepwise reasoning.

\paragraph{LogiQA}
LogiQA~\cite{liu2020logiqa} is a dataset crafted specifically to test logical reasoning in reading comprehension, containing questions that demand deductive and inductive inference. Models must analyze logical structures in text, making it a valuable resource for code-enhanced techniques that incorporate symbolic reasoning or rule-based algorithms. Success on LogiQA requires coherent, step-by-step thinking and the ability to identify logical entailments.

\paragraph{DROP}
DROP~\cite{dua2019drop} challenges models to perform numerical and symbolic manipulations to answer questions. It often involves arithmetic operations, entity tracking, and multi-step logic derivations, making it an excellent testbed for code-driven reasoning strategies. By leveraging program-like steps to parse text and compute answers, models can demonstrate deeper reasoning skills.

\paragraph{MathShepherd-pair}
MathShepherd-pair~\cite{wang2024math} focuses on pairwise comparisons of mathematical reasoning steps, often requiring validation of correctness or logical consistency. It encourages the use of code-like procedures—such as symbolic manipulation or step-by-step solution checking—to ensure precise, verifiable reasoning. This pairing format helps evaluate a model’s ability to systematically analyze and contrast different solution paths.

\paragraph{ReClor-pair}
ReClor-pair~\cite{yu2020reclor} extends the ReClor dataset’s focus on complex logical reasoning by providing question-answer pairs that examine a model’s capacity for distinguishing subtle logical cues. The paired setup highlights the necessity of structured, often code-driven verification mechanisms, where models benefit from systematically comparing and validating reasoning options. Performance here is indicative of robust logical inference capabilities.

\paragraph{LogiQA2.0-pair}
LogiQA2.0-pair~\cite{liu2023logiqa} offers an updated set of logical reasoning challenges in a paired format, demanding thorough analysis of propositions and argument structures. By encouraging code-enhanced methods—like building parse trees or applying logical inference rules—this dataset underscores the importance of systematic step-by-step reasoning. It is particularly useful for benchmarking improvements in logical rigor.

\paragraph{APE}
APE~\cite{zhao2020ape210k} tasks revolve around interpreting arithmetic or algorithmic steps and providing a rationale. Models trained with code are better positioned to explain or verify each step programmatically. The dataset pushes for explanatory reasoning, where each numeric or logical action needs to be justified systematically.

\paragraph{CMATH}
CMATH~\cite{wei2023cmath} contains math problems, typically in a non-English (e.g., Chinese) context, testing a model’s ability to parse language-specific nuances and generate reasoned steps. Its design demands clear logical structuring, often improved by programmatic solution paths that systematically handle textual variations. Code-enhanced methods help unify language understanding with algorithmic resolution of math tasks.

\paragraph{AlpacaEval-2}
AlpacaEval-2~\cite{li2023alpacaeval} is an instruction-following evaluation suite that includes tasks requiring reasoning and structured thinking. While not exclusively code-based, the dataset benefits from code-infused methods that guide stepwise logic, especially for tasks involving multi-turn reasoning or systematic dissection of instructions. It thus measures how effectively models integrate reasoning processes into instruction comprehension.

\paragraph{MT-Bench}
MT-Bench~\cite{zheng2023judging} is a multi-turn benchmark that assesses conversational coherence, reasoning depth, and consistency over extended dialogues. It tests whether models can maintain logical continuity and sound reasoning across multiple exchanges. Code-centric approaches—such as planning-based or programmatic reasoning—can boost the clarity and correctness of the model’s dialogue responses.

\paragraph{ALFWorld}
ALFWorld~\cite{shridhar2020alfworld} places agents in interactive text-based environments that require sequential decision-making and reasoning about cause-and-effect. Models must combine language understanding with environmental cues to perform complex tasks, often using reasoning strategies resembling small programs or scripts. This environment underscores the importance of code-level logic for planning and executing multi-step goals.

\subsection{Reasoning-enhanced Code Intelligence}

\noindent
The development of robust code intelligence systems necessitates comprehensive evaluation frameworks. This section presents key benchmarks that assess various aspects of code generation and understanding, ranging from functional understanding and correctness and algorithmic problem-solving to repository-level understanding modifications. These benchmarks provide standardized metrics for measuring progress in code intelligence, with particular emphasis on real-world applicability and systematic evaluation of reasoning capabilities in programming contexts.

\paragraph{HumanEval}
HumanEval~\cite{chen2021evaluatinglargelanguagemodels} evaluates the functional correctness of code generated by large language models by presenting 164 hand-crafted programming challenges. Each problem is defined by a function signature, a descriptive docstring, and a set of unit tests (averaging around 7.7 tests per problem), which together verify that the generated solution meets the intended functionality via the pass@k metric. This benchmark primarily focuses on assessing models’ ability to translate natural language prompts into functionally correct code.

\paragraph{MBPP}
MBPP~\cite{austin2021programsynthesislargelanguage} comprises approximately 1,000 Python programming problems that pair natural language descriptions with corresponding code solutions and multiple automated test cases. By measuring whether the generated code passes these tests, MBPP benchmarks models on their capability to synthesize accurate and executable Python code from plain language instructions, emphasizing fundamental programming skills and effective problem decomposition.

\paragraph{APPS}
APPS~\cite{hendrycks2021measuringcodingchallengecompetence} provides a diverse evaluation framework consisting of around 10,000 problems, ranging from simple one-line solutions to complex algorithmic challenges. The benchmark employs unit tests to determine the functional correctness of generated code, thereby benchmarking the models on their versatility and ability to handle a broad spectrum of programming scenarios under realistic conditions.

\paragraph{DS-1000}
DS-1000~\cite{lai2022ds1000naturalreliablebenchmark} is a specialized benchmark tailored to the data science domain, focusing on code generation tasks that involve data manipulation, statistical analysis, and data visualization. By incorporating challenges that demand domain-specific knowledge and practical data-handling skills, DS-1000 uniquely evaluates a model’s ability to produce contextually relevant and functionally correct code for data-centric applications.

\paragraph{RepoBench}
RepoBench~\cite{liu2023repobench} is a benchmark specifically designed for evaluating repository-level code auto-completion systems. Its abstract outlines three interlinked evaluation tasks—RepoBench-R (Retrieval), RepoBench-C (Code Completion), and RepoBench-P (Pipeline)—which collectively assess a system’s ability to extract relevant cross-file code snippets, integrate both in-file and cross-file contexts, and predict the next line of code in complex, multi-file programming scenarios. This approach fills the gap left by prior single-file benchmarks and facilitates a comprehensive comparison of auto-completion performance.

\paragraph{CrossCodeEval}
CrossCodeEval~\cite{ding2023crosscodeevaldiversemultilingualbenchmark} presents a diverse and multilingual benchmark that targets the challenges of cross-file code completion. According to its abstract, the benchmark is built on real-world, open-sourced repositories in four popular programming languages—Python, Java, TypeScript, and C\#—and features examples that strictly require leveraging information from multiple files for accurate code completion. The work emphasizes a static-analysis-based method to pinpoint instances where cross-file context is essential, thereby evaluating both code generation and context retrieval capabilities under realistic conditions.

\paragraph{LiveCodeBench}
LiveCodeBench~\cite{jain2024livecodebenchholisticcontaminationfree} is a holistic, contamination‐free evaluation benchmark for code, continuously collecting new, high-quality coding problems over time from LeetCode, AtCoder, and CodeForces. It extends traditional evaluation by incorporating not only code generation but also broader code-related capabilities such as self‐repair, execution, and test output prediction. By using a time-sensitive collection of challenges, LiveCodeBench aims to assess models on truly unseen problems, ensuring that performance measurements remain robust and reflective of real-world development scenarios.

\paragraph{BigCodeBench}
BigCodeBench~\cite{zhuo2024bigcodebench} is a comprehensive benchmark for assessing large-scale code generation and understanding, which encompasses a wide variety of programming languages and repository complexities, challenging models with real-world coding scenarios that include intricate multi-file dependencies and extensive project structures. Designed to stress-test model capabilities on both functional correctness and code synthesis quality, BigCodeBench provides a scalable evaluation framework that mirrors the heterogeneity encountered in open-source codebases.

\paragraph{CRUXEval} CRUXEval~\cite{gu2024cruxevalbenchmarkcodereasoning} is a benchmark containing 800 short Python functions, ranging from 3 to 13 lines, each paired with input-output examples. It defines two tasks: input prediction for evaluating code reasoning and understanding, and output prediction for assessing execution behavior. 

\paragraph{RepoQA} RepoQA~\cite{liu2024repoqaevaluatinglongcontext} is a benchmark designed to evaluate long-context code understanding through realistic codebase search scenarios. It consists of 500 code search tasks drawn from 50 popular repositories across five programming languages. Using a "needle-in-a-haystack" approach, models must locate specific code snippets within extensive contextual code. The benchmark evaluates both retrieval accuracy and comprehension of multi-file, long-context code environments, reflecting real-world developer challenges.

\paragraph{SWE-bench}
SWE-bench~\cite{jimenez2024swebenchlanguagemodelsresolve} s a software engineering benchmark based on real GitHub issues and corresponding pull requests. Each evaluation task requires generating a fix patch in complex, multi-file repositories to resolve specific issues. The evaluation system uses the repository's original unit testing framework to verify the correctness of solutions. By simulating challenges encountered in actual software development, SWE-bench provides a realistic evaluation environment.

\paragraph{SWE-bench Multimodal} 
SWE-bench Multimodal~\cite{yang2024swebenchmultimodalaisystems} extends SWE-bench by incorporating visual inputs. The dataset is collected from JavaScript repositories, where each task instance includes images embedded in problem descriptions or unit tests, focusing on front-end development areas like UI design, diagramming, and data visualization. This benchmark evaluates AI systems' ability to generalize across different modalities and programming paradigms by integrating visual elements.

\paragraph{SWE-bench Verified} 
SWE-bench Verified~\cite{chowdhury2024swebenchverified} is an optimized version of SWE-bench containing a human-validated subset. Developers rigorously annotated and screened task instances to remove underspecified or ambiguous cases. Each instance contains reliable ``fail-to-pass" unit tests and clear issue descriptions, providing a more accurate measure of a model's capability to resolve real-world software issues.

\section{Paper Collection}

To ensure comprehensive coverage of relevant literature, we employed a systematic paper collection approach. We utilized arXiv as our primary source and conducted searches using a combination of keywords: ("code" OR "program") AND ("reason" OR "plan"). We restricted our search to papers within the Computer Science - Artificial Intelligence (cs.AI) and Computer Science - Computation and Language (cs.CL) categories, focusing on works published after January 2021. This timeframe was chosen deliberately as it marks a significant turning point in code reasoning research, coinciding with the emergence of large language models like Codex and the subsequent surge in research combining natural language processing with code understanding. Our initial search yielded 110 papers. Subsequently, we performed a manual filtering process, carefully examining each paper's relevance, technical depth, and contributions to the field of code reasoning. This thorough inspection resulted in a final collection of 63 papers that form the core of our survey. These selected papers represent the most significant and relevant contributions to understanding the interplay between code and reasoning in recent years.

\section{Additional Tables and Figures}

\begin{table*}[ht!]
\centering
\resizebox{\textwidth}{!}{

\begin{tabular}{@{}l l l c c c c c c c c c@{}}
\toprule
\textbf{Method} & \textbf{Model} & \textbf{Settings} & \textbf{GSM8K} & \textbf{GSM-HARD} & \textbf{SVAMP} & \textbf{ASDiv} & \textbf{SingleEq} & \textbf{AddSub} & \textbf{MultiArith} & \textbf{MATH} & \textbf{AQuA}\\
\midrule

% DIRECT
\multirow{1}{*}{\textbf{Direct$^\dagger$}}
& Codex & Few-shot Direct Prompting
& 19.7 & 5.0 & 69.9 & 74.0 & 86.8 & 90.9 & 44.0 & -- & -- \\
\midrule

% CoT
\multirow{10}{*}{\textbf{CoT$^\dagger$}~\cite{wei2022chain}}
& UL2-20B      & Few-shot Chain-of-Thought       & 4.1  & --   & 12.6 & 16.9  & --   & 18.2 & 10.7 & --   & -- \\
& LaMDA-137B   & Few-shot Chain-of-Thought       & 17.1 & --   & 39.9 & 49.0  & --   & 52.9 & 51.8 & --   & -- \\
& Codex        & Few-shot Chain-of-Thought       & 65.6 & 23.1 & 74.8 & 76.9  & 89.1 & 86.0 & 95.9 & --   & -- \\
& PaLM-540B    & Few-shot Chain-of-Thought       & 56.9 & --   & 79.0 & 73.9  & 92.3 & 91.9 & 94.7 & --   & -- \\
& Minerva-540B & Few-shot Chain-of-Thought       & 58.8 & --   & --   & --    & --   & --   & --   & --   & -- \\
& GPT-4        & Few-shot Chain-of-Thought       & 92.0 & --   & 97.0 & --    & --   & --   & --   & --   & -- \\
& GPT-4o-mini  & 0-shot Chain-of-Thought         & --   & --   & --   & --    & --   & --   & --   & 50.6 & -- \\
& Llama3.1-8B  & 0-shot Chain-of-Thought         & --   & --   & --   & --    & --   & --   & --   & 18.3 & -- \\
& GPT-3.5      & 0-shot Chain-of-Thought         & 81.6 & --   & 78.2 & --    & 93.1 & 86.1 & 96.7 & --   & -- \\
& GPT-3.5      & Few-shot Chain-of-Thought       & 82.1 & --   & 77.1 & --    & 95.5 & 90.6 & 98.5 & --   & -- \\
\midrule

% PAL
\multirow{4}{*}{\textbf{PAL}~\cite{kabra2023program}}
& Codex        & Few-shot Program-aided LM       & 72.0 & 61.2 & 79.4 & 79.6 & 96.1 & 92.5 & 99.2 & --   & -- \\
& GPT-4o-mini  & 0-shot Program-aided LM         & --   & --   & --   & --   & --   & --   & --   & 36.6 & -- \\
& Llama3.1-8B  & 0-shot Program-aided LM         & --   & --   & --   & --   & --   & --   & --   & 11.7 & -- \\
& GPT-3.5      & Few-shot Program-aided LM       & 80.6 & --   & 79.5 & --   & 97.6 & 89.1 & 97.0 & --   & -- \\
\midrule

% PoT
\multirow{3}{*}{\textbf{PoT}~\cite{chen2022program}}
& Codex        & Few-shot Program of Thought                      & 71.6 & --   & 85.2 & --   & --   & --   & --   & 54.1 & 54.1 \\
& Codex        & Few-shot Program of Thought + Self-Consistency   & 80.0 & --   & 89.1 & --   & --   & --   & --   & --   & --   \\
& GPT-4        & Few-shot Program of Thought                      & 97.2 & --   & 97.4 & --   & --   & --   & --   & --   & --   \\
\midrule

% MathCoder-L
\multirow{3}{*}{\textbf{MathCoder}~\cite{wang2023mathcoder}}
& Llama-2-7B   & 0-shot Code Interleaving / Fine-tuned            & 64.2 & --   & 71.5 & --   & --   & --   & --   & 23.3 & -- \\
& Llama-2-13B  & 0-shot Code Interleaving                         & 72.6 & --   & 76.9 & --   & --   & --   & --   & 29.9 & -- \\
& Llama-2-70B  & 0-shot Code Interleaving                         & 83.9 & --   & 84.9 & --   & --   & --   & --   & 45.1 & -- \\
\midrule

% MathCoder-CL
\multirow{3}{*}{\textbf{MathCoder2}~\cite{lu2024mathcoder2}}
& CodeLlama-7B  & 0-shot Code Interleaving                        & 67.8 & --   & 70.7 & --   & --   & --   & --   & 30.2 & -- \\
& CodeLlama-13B & 0-shot Code Interleaving                        & 74.1 & --   & 78.0 & --   & --   & --   & --   & 35.9 & -- \\
& CodeLlama-34B & 0-shot Code Interleaving                        & 81.7 & --   & 82.5 & --   & --   & --   & --   & 45.2 & -- \\
\midrule

% CodePlan
\multirow{3}{*}{\textbf{CodePlan}~\cite{wen2024unlocking}}
& Mistral-7B    & Few-shot Code-form planning                     & 59.5 & --   & 61.4 & --   & --   & --   & --   & 34.3 & -- \\
& Llama-2-7B    & Few-shot Code-form planning                     & 33.8 & --   & 41.5 & --   & --   & --   & --   & 20.8 & -- \\
& Llama-2-13B   & Few-shot Code-form planning                     & 49.5 & --   & 53.4 & --   & --   & --   & --   & 27.4 & -- \\
\midrule

% % CodeNL
% \multirow{2}{*}{\textbf{CodeNL}}
% & GPT-4o-mini   & 0-shot Code Prompting                           & --   & --   & --   & --   & --   & --   & --   & 50.9 & -- \\
% & Llama3.1-8B   & 0-shot Code Prompting                           & --   & --   & --   & --   & --   & --   & --   & 14.9 & -- \\
% \midrule

% % NLCode
% \multirow{2}{*}{\textbf{NLCode}}
% & GPT-4o-mini   & 0-shot Code Prompting                           & --   & --   & --   & --   & --   & --   & --   & 47.6 & -- \\
% & Llama3.1-8B   & 0-shot Code Prompting                           & --   & --   & --   & --   & --   & --   & --   & 18.4 & -- \\
% \midrule

% INC-Math
\multirow{2}{*}{\textbf{INC-Math}~\cite{xiong2024inc}}
& GPT-4o-mini   & 0-shot Code Prompting                           & --   & --   & --   & --   & --   & --   & --   & 51.4 & -- \\
& Llama3.1-8B   & 0-shot Code Prompting                           & --   & --   & --   & --   & --   & --   & --   & 16.7 & -- \\
\midrule

% CoC
\multirow{1}{*}{\textbf{CoC}~\cite{li2023chain}}
& text-davinci-003 & Few-shot Code Interleaving with Python Exec. & 71.0 & --   & --   & --   & --   & --   & --   & --   & -- \\
\midrule

% Code
\multirow{2}{*}{\textbf{CodePrompt}~\cite{hu2023code}}
& GPT-3.5 & 0-shot Code Prompting with self-debug
& 78.9 & -- & 79.4 & -- & 97.6 & 91.7 & 96.7 & -- & -- \\
& GPT-3.5 & Few-shot Code Prompting with self-debug
& 80.6 & -- & 79.6 & -- & 97.4 & 91.4 & 97.3 & -- & -- \\
\bottomrule
\end{tabular}}
\caption{Performance of various code-aided reasoning methods on multiple benchmarks. ``--'' indicates no reported result. Numerical results represent the percentage of problems that were solved correctly. $^\dagger$ Direct and CoT results are from~\citet{chen2022program}.}
\label{tab:code_reasoning_methods}
\end{table*}

\begin{table*}[ht!]
\centering
\resizebox{\textwidth}{!}{
\begin{tabular}{@{}lllcccc@{}}
\toprule
\textbf{Method} & \textbf{Model} & \textbf{Settings} & \textbf{HumanEval} & \textbf{MBPP} & \textbf{SWE-Bench (Lite) } \\
\midrule

% DIRECT
\multirow{14}{*}{\textbf{Direct}$^{\dagger}$ }
& AlphaCode-1.1B & 0-shot Prompting & 17.1 & -- & -- \\
& Incoder-6.7B & 0-shot Prompting & 15.2 & 17.6 & -- \\
& CodeGeeX-13B & 0-shot Prompting & 18.9 & 26.9 & -- \\
& StarCoder-15.5B & 0-shot Prompting & 34.1 & 43.6 & -- \\
& CodeLlama-34B & 0-shot Prompting & 51.8 & 69.3 & -- \\
& Llama3-8B & 0-shot Prompting & 62.2 & -- & -- \\
& CodeGen-Mono-16.1B & 0-shot Prompting & 32.9 & 38.6 & -- \\
& Codex & 0-shot Prompting & 47.0 & 58.1 & -- \\
& Codex+CodeT & 0-shot Prompting & 65.8 & 67.7 & -- \\
& GPT-3.5 Turbo & 0-shot Prompting & 57.3 & 52.2 & -- \\
& PaLM Coder & 0-shot Prompting & 43.9 & 32.3 & -- \\
& Claude-instant-1 & 0-shot Prompting & 31.1 & 26.9 & -- \\
& GPT-4 Turbo & 0-shot Prompting & 57.9 & 63.4 & -- \\
& GPT-4 & 0-shot Prompting & 67.6 & 68.3 & -- \\
\midrule

% CoT
\multirow{2}{*}{\textbf{CoT}~\cite{wei2023chainofthoughtpromptingelicitsreasoning}} 
& GPT-3.5$^{\ddagger}$ & 0-shot Chain-of-Thought & 44.6 & 46.1 & -- \\
& Codex$^{\star}$ & Few-shot Chain-of-Thought  & 53.9   & 54.5  & --   \\
\midrule

% Self-Edit
\multirow{3}{*}{\textbf{Self-Edit}~\cite{zhang2023selfeditfaultawarecodeeditor}} 
& InCoder-1B & 0-shot Prompting & 3.7 & -- & -- \\
& CodeGen-2B & 0-shot Prompting & 17.1 & -- & -- \\
& GPT-3 & Few-shot Prompting   & 39.6    & --    & --   \\
\midrule

% Self-Planning
\multirow{3}{*}{\textbf{Self-Planning}~\cite{jiang2024selfplanningcodegenerationlarge}} 
& Codex        & Few-shot Prompting & 60.3  & 55.7  & --    \\ 
& text-davinci-003 & Few-shot Prompting & 65.4 & -- & - \\
& GPT-3 & Few-shot Prompting &  50.0 & -- & -- \\
\midrule

% Self-Debugging
\multirow{4}{*}{\textbf{Self-Debugging}~\cite{chen2023teachinglargelanguagemodels}} 
& StarCoder &  Few-shot Prompting & -- & 53.2 & -- \\
& Codex  & Few-shot Prompting & --  & 70.8  & --    \\
& GPT-3.5  & Few-shot Prompting   & --    & 74.2    &    \\
& GPT-4  & Few-shot Prompting  & --    & 80.6    & --    \\
\midrule

% Self-Collaboration
\multirow{1}{*}{\textbf{Self-Collaboration}~\cite{dong2024selfcollaborationcodegenerationchatgpt}} 
& GPT-3.5  & Few-shot Prompting   & 74.4    & 68.2    &    \\
\midrule

% SCoTs
\multirow{2}{*}{\textbf{SCoTs}~\cite{li2023structuredchainofthoughtpromptingcode}} 
& Codex        & Few-shot Prompting & 49.8  & 38.3  & --    \\ 
& GPT-3.5 & Few-shot Prompting   & 60.6    & 47.0    & --   \\
\midrule

% CodeCoT
\multirow{1}{*}{\textbf{CodeCoT}~\cite{huang2024codecottacklingcodesyntax}} 
& GPT-3.5 & Few-shot Prompting   & 79.3    & 89.5    & --   \\
\midrule

% CodeAct
\multirow{2}{*}{\textbf{CodeAct}~\cite{wang2024executablecodeactionselicit}} 
& Llama2-7B        &  Fine-tuning & 18.1  & -- & --    \\
& Mistral-7B     &  Fine-tuning & 34.7  & --  & --    \\
\midrule

% OpenCodeInterpreter
\multirow{8}{*}{\textbf{OpenCodeInterpreter}$^{\P}$~\cite{zheng2025opencodeinterpreterintegratingcodegeneration}} 
& CodeLlama-Python-7B & Fine-tuning & 75.6 & 69.9 & -- \\
& StarCoder2-7B & Fine-tuning & 75.6 & 66.9 & -- \\
& DeepseekCoder-6.7B & Fine-tuning & 81.1 & 82.7 & -- \\
& StarCoder2-15B & Fine-tuning & 77.4 & 74.2 & -- \\
& CodeLlama-Python-13B & Fine-tuning & 81.1 & 78.2 & -- \\
& CodeLlama-Python-34B & Fine-tuning & 81.7 & 80.2 & -- \\
& DeepseekCoder-33B & Fine-tuning & 82.9 & 83.5 & -- \\
& CodeLlama-Python-70B & Fine-tuning & 79.9 & 81.5 & -- \\
\midrule

% AgentCoder
\multirow{5}{*}{\textbf{AgentCoder}~\cite{zhang2024codeagentenhancingcodegeneration}} 
& GPT-3.5 Turbo & Agentic Prompting & 79.9 & 89.9 & -- \\
& PaLM Coder & Agentic Prompting & 64.0 & 75.9 & -- \\
& Claude-instant-1 & Agentic Prompting & 67.7 & 76.3 & -- \\
& GPT-4 & Agentic Prompting & 96.3 & 91.8 & -- \\
& GPT-4 Turbo & Agentic Prompting & 89.6 & 91.4 & -- \\
\midrule

% SWE-agent
\multirow{4}{*}{\textbf{SWE-agent}~\cite{yang2024sweagentagentcomputerinterfacesenable}} 
& Claude 3 Opus & Agentic Prompting & -- & -- & 13.0 \\
& GPT-4 Turbo & Agentic Prompting & -- & -- & 18.0 \\
& Claude 3.5 Sonnet$^{\diamond}$  & Agentic Prompting & -- & -- & 23.0 \\
& Claude 3.5 Sonnet + o1$^{\diamond}$  & Agentic Prompting & -- & -- & 45.3 \\
\midrule

% Agentless
\multirow{1}{*}{\textbf{Agentless}~\cite{xia2024agentlessdemystifyingllmbasedsoftware}} 
& GPT-4o & Agentic Prompting & -- & -- & 27.3 \\
\midrule

% OpenHands
\multirow{3}{*}{\textbf{OpenHands}~\cite{openhands2024}} 
& GPT-4o-mini & Agentic Prompting & -- & -- & 6.3 \\
& GPT-4o & Agentic Prompting & -- & -- & 22.0 \\
& Claude 3.5 Sonnet & Agentic Prompting & -- & -- & 26.0 \\
\midrule

% AutoCodeRover
\multirow{2}{*}{\textbf{AutoCodeRover}~\cite{zhang2024autocoderoverautonomousprogramimprovement}} 
& GPT-4 & Agentic Prompting & -- & -- & 19.0 \\
& GPT-4o$^{\bullet}$ & Agentic Prompting & -- & -- & 22.7 \\
\midrule

% HyperAgent
\multirow{1}{*}{\textbf{HyperAgent}~\cite{phan2024hyperagentgeneralistsoftwareengineering}} 
& Claude-3.5-Sonnet & Agentic Prompting & -- & -- & 26.0 \\

\bottomrule
\end{tabular}
}
\caption{Performance of various reasoning-enhanced code intelligence methods on multiple benchmarks. Results from original papers unless noted otherwise. HumanEval and MBPP use pass@1 scoring. $^{\dagger}$Results for all Direct methods are from the AgentCoder paper~\cite{huang2023agentcoder}. $^{\ddagger}$Result from Self-Collaboration paper~\cite{dong2024selfcollaborationcodegenerationchatgpt}. $^{\star}$Result from Self-Planning paper~\cite{jiang2024selfplanningcodegenerationlarge}. $^{\P}$We report the results with execution feedback (but without human involvement). $^{\diamond}$Results from official SWE-bench leaderboard (accessed Feb 15, 2025). $^{\bullet}$Result from HyperAgent paper~\cite{phan2024hyperagentgeneralistsoftwareengineering}.}
\label{tab:full_reasoning_code_methods}
\end{table*}

\definecolor{hidden-pink}{RGB}{255,245,247}
\definecolor{hidden-draw}{RGB}{107,114,128}

\definecolor{level-1}{RGB}{240,249,255}  % Light blue
\definecolor{level-2}{RGB}{236,253,245}  % Light green
\definecolor{level-3}{RGB}{254,242,242}  % Light red

\tikzstyle{my-box}=[
    rectangle,
    draw=hidden-draw,
    rounded corners,
    text opacity=1,
    minimum height=2em,
    minimum width=5em,
    inner sep=2pt,
    align=center,
    fill opacity=.5,
    line width=0.8pt,
]
\tikzstyle{citation}=[my-box, minimum height=2em,
    fill=hidden-pink!80, text=black, align=left,font=\normalsize,
    inner xsep=2pt,
    inner ysep=5pt,
    line width=0.8pt,
]
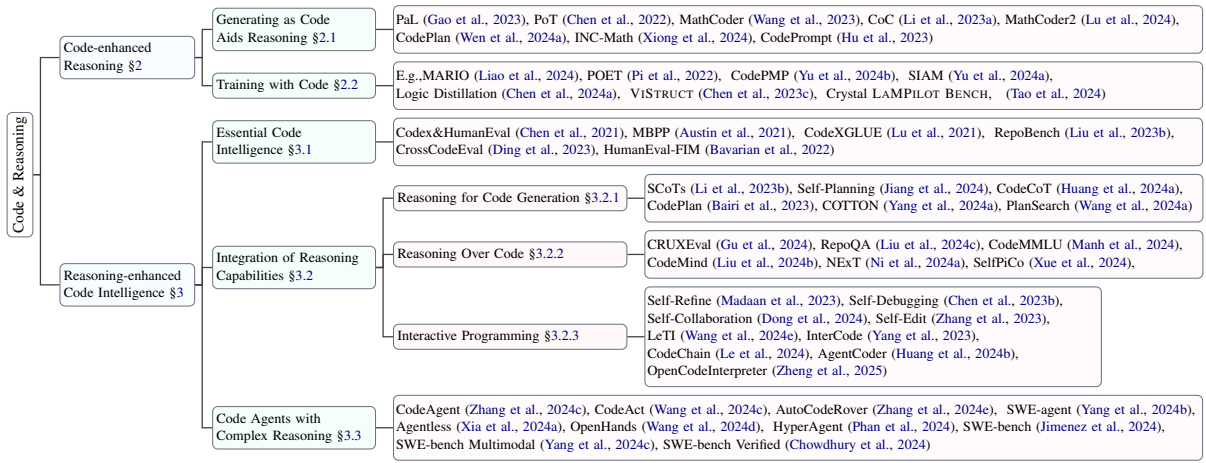
\begin{figure*}[!th]
    \centering
    \resizebox{0.99\textwidth}{!}{
        \begin{forest}
            forked edges,
            for tree={
                grow=east,
                reversed=true,
                anchor=base west,
                parent anchor=east,
                child anchor=west,
                base=left,
                font=\large,
                rectangle,
                draw=hidden-draw,
                rounded corners,
                align=left,
                minimum width=6em,
                edge+={darkgray, line width=1pt},
                s sep=6pt,
                inner xsep=3pt,
                inner ysep=3pt,
                line width=0.8pt,
                ver/.style={rotate=90, child anchor=north, parent anchor=south, anchor=center},
            },
            where level=1{
                text width=9.0em,
                font=\normalsize,
                fill=level-1,
                fill opacity=.5,
            }{},
            where level=2{
                text width=11.0em,
                font=\normalsize,
                fill=level-2,
                fill opacity=.5,
            }{},
            where level=3{
                text width=16.0em,
                font=\normalsize,
                fill=level-3,
                fill opacity=.5,
            }{},
            where level=4{
                text width=18.0em,
                font=\normalsize,
                fill=hidden-pink,
                fill opacity=.5,
            }{}
            [
                Code \& Reasoning , ver
                [                
                    Code-enhanced \\Reasoning \S\ref{sec:cer}
                    [
                        Generating as Code\\Aids Reasoning \S\ref{sec:c4r}
                        [
                            PaL~\cite{gao2023pal}{, }PoT~\cite{chen2022program}{, }MathCoder~\cite{wang2023mathcoder}{, }CoC~\cite{li2023chain}{, }MathCoder2~\cite{lu2024mathcoder2}{, }\\CodePlan~\cite{wen2024unlocking}{, }INC-Math~\cite{xiong2024inc}{, }CodePrompt~\cite{hu2023code}
                            , citation, text width=57em
                        ]
                    ]
                    [
                        Training with Code \S\ref{sec:twc}
                        [
                            {E.g.,}MARIO~\cite{liao2024mario}{, }POET~\cite{pi2022reasoning}{, }
                            CodePMP~\cite{yu2024codepmp}{, }
                            SIAM~\cite{yu2024siam}{, } \\
                            Logic Distillation~\cite{chen2024logic}{, }
                            \textsc{ViStruct}~\citep{chen2023vistruct}{, }
                            Crystal~\textsc{LaMPilot Bench}{, }
                            ~\cite{tao2024crystal}
                            , citation, text width=57em
                        ]
                    ]
                ]
                [
                    Reasoning-enhanced \\Code Intelligence \S\ref{sec:rec}
                    [
                        Essential Code \\ Intelligence \S\ref{sec:eci}
                        [
                            Codex\&HumanEval~\cite{chen2021evaluatinglargelanguagemodels}{, }MBPP~\cite{austin2021programsynthesislargelanguage}{, } CodeXGLUE~\cite{lu2021codexgluemachinelearningbenchmark}{, }
                            RepoBench~\cite{liu2023repobenchbenchmarkingrepositorylevelcode}{, } \\ CrossCodeEval~\cite{ding2023crosscodeevaldiversemultilingualbenchmark}{, }HumanEval-FIM~\cite{bavarian2022efficienttraininglanguagemodels}
                            , citation, text width=57em
                        ]
                    ]
                    [
                        Integration of Reasoning \\Capabilities \S\ref{sec:irc}
                        [
                            Reasoning for Code Generation \S\ref{sec:r4c}
                            [
                                SCoTs~\cite{li2023structuredchainofthoughtpromptingcode}{, }Self-Planning~\cite{jiang2024selfplanningcodegenerationlarge}{, }CodeCoT~\cite{huang2024codecottacklingcodesyntax}{, } \\CodePlan~\cite{bairi2023codeplanrepositorylevelcodingusing}{, }COTTON~\cite{yang2024chainofthoughtneuralcodegeneration}{, }PlanSearch~\cite{wang2024planningnaturallanguageimproves}
                                , citation, text width=39.2em
                            ]
                        ]
                        [
                            Reasoning Over Code \S\ref{sec:roc}
                            [
                                CRUXEval~\cite{gu2024cruxevalbenchmarkcodereasoning}{, }RepoQA~\cite{liu2024repoqaevaluatinglongcontext}{, }CodeMMLU~\cite{manh2024codemmlumultitaskbenchmarkassessing}{, } \\
                                CodeMind~\cite{liu2024codemindframeworkchallengelarge}{, }NExT~\cite{ni2024nextteachinglargelanguage}{, }SelfPiCo~\cite{xue2024selfpicoselfguidedpartialcode}{, }, citation, text width=39.2em
                            ]
                        ]
                        [
                            Interactive Programming \S\ref{sec:ip}
                            [
                                Self-Refine~\cite{madaan2023selfrefineiterativerefinementselffeedback}{, }Self-Debugging~\cite{chen2023teachinglargelanguagemodels}{, }\\
                                Self-Collaboration~\cite{dong2024selfcollaborationcodegenerationchatgpt}{, }Self-Edit~\cite{zhang2023selfeditfaultawarecodeeditor}{, }\\
                                LeTI~\cite{wang2024letilearninggeneratetextual}{, }InterCode~\cite{yang2023intercodestandardizingbenchmarkinginteractive}{, }\\
                                CodeChain~\cite{le2024codechainmodularcodegeneration}{, }AgentCoder~\cite{huang2024agentcodermultiagentbasedcodegeneration}{, }\\
                                OpenCodeInterpreter~\cite{zheng2025opencodeinterpreterintegratingcodegeneration}
                                , citation, text width=32em
                            ]
                        ]
                    ]
                    [
                        Code Agents with \\Complex Reasoning \S\ref{sec:agent}
                        [
                            CodeAgent~\cite{zhang2024codeagentenhancingcodegeneration}{, }CodeAct~\cite{wang2024executablecodeactionselicit}{, }AutoCodeRover~\cite{zhang2024autocoderoverautonomousprogramimprovement}{, }
                            SWE-agent~\cite{yang2024sweagentagentcomputerinterfacesenable}{, }\\Agentless~\cite{xia2024agentlessdemystifyingllmbasedsoftware}{, }OpenHands~\cite{openhands2024}{, }
                            HyperAgent~\cite{phan2024hyperagentgeneralistsoftwareengineering}{, }SWE-bench~\cite{jimenez2024swebenchlanguagemodelsresolve}{, }\\
                            SWE-bench Multimodal~\cite{yang2024swebenchmultimodalaisystems}{, }SWE-bench Verified~\cite{chowdhury2024swebenchverified}
                            , citation, text width=57em
                        ]
                    ]
                ]
            ]
        \end{forest}
    }
    \caption{Full taxonomy illustrating the interplay between code and reasoning.}
    \label{fig:taxonomy_full}
\end{figure*}

\end{document}